%% file: main.tex
\documentclass[runningheads]{llncs}

\usepackage[T1]{fontenc}
\usepackage{microtype}
\usepackage{amsmath,amssymb}
\let\origRightarrow\Rightarrow
\usepackage{marvosym}
\let\Rightarrow\origRightarrow
\usepackage{bm}
\usepackage{graphicx}
\usepackage{booktabs}
\usepackage{multirow}
\usepackage{makecell}
\usepackage{array}
\usepackage[section]{placeins}
\usepackage[table]{xcolor}
\usepackage{url}
\usepackage[hidelinks,
  pdftitle={BRACE: Calibrated Multi-View Tabular Ensembles for Interbank Credit Rating and Graph Anomaly Detection},
  pdfauthor={Zihe Wei, Zilong Yang, and Sheng Xiang}
]{hyperref}

\newcommand{\best}[1]{\textbf{#1}}
\newcommand{\runnerup}[1]{\underline{#1}}
\newcommand{\corr}{\textsuperscript{(\Letter)}}

\begin{document}

\title{BRACE: Calibrated Multi-View Tabular Ensembles for Interbank Credit Rating and Graph Anomaly Detection}
\titlerunning{BRACE: Calibrated Multi-View Tabular Ensembles}

\author{Zihe Wei \and Zilong Yang \and Sheng Xiang\corr}
\authorrunning{Z. Wei et al.}
\institute{School of Computer Science and Technology, Tongji University, Shanghai, China\\
\email{\{weizihe, 2353257, xiangsheng\}@tongji.edu.cn}}

\maketitle

\input{sections/00_abstract}
\input{sections/01_intro}
\input{sections/02_problem}
\input{sections/03_method}
\input{sections/04_setup}
\input{sections/05_results}
\input{sections/06_related}
\input{sections/07_conclusion}

\bibliographystyle{splncs04}
\bibliography{refs}

\input{sections/08_appendix}

\end{document}

%% file: sections/00_abstract.tex
\begin{abstract}
Risk assessment on networked entities is essential for credit rating and graph anomaly detection. Traditional tabular approaches focus on isolated node features, overlooking historical transitions and relational context. Graph-based techniques, particularly Graph Neural Networks (GNNs), advance node-level modeling using relational information. However, in real-world benchmarks, topologies span reconstructed bank exposures and heterophilous, disconnected anomaly graphs under shifting priors. Existing GNN and graph-flattening methods, though effective in some scenarios, struggle to separate informative edges from noise. To address these challenges, we propose a novel framework, Bank Rating with Calibrated Ensembles (BRACE), decoupling views from calibrated decisions. Specifically, our approach pairs tabular foundation priors with convex log-score pooling under uncertain graph quality. Banking panels are first converted into ego and historical-transition views, whereas static anomaly graphs form ego, relational, and deflated Beta wavelet views in \emph{BRACE-Ultra}. Then we cross-fit tabular foundation models, plus cutoff-realigned historical heads on panels, gating views via KKT-certified log-score pooling. Furthermore, bias-corrected temperature scaling, panel MAP-EM, and F1-aligned decoding calibrate imbalanced decisions. Experiments on interbank rating and four anomaly graphs demonstrate that our approach outperforms state-of-the-art models. Additionally, U.S.\ Call Report transfer and controlled edge-rewiring tests validate BRACE's robustness and practical applicability. Code: \url{https://github.com/AI4DataSynth/BRACE}.
\keywords{Credit rating \and Tabular foundation models \and Spectral wavelets \and Calibration \and Graph anomaly detection.}
\end{abstract}

%% file: sections/01_intro.tex
\section{Introduction}
\label{sec:intro}

Systemic credit defaults and organized digital fraud inflict severe economic losses, drawing continuous attention from financial institutions, platform operators, and regulators. Reliable risk assessment across networked entities is essential for safeguarding counterparty solvency, pricing credit, and curbing anomalies under uneven disclosures and uncertain relational links.

\begin{figure}[t]
\centering
\includegraphics[width=\textwidth]{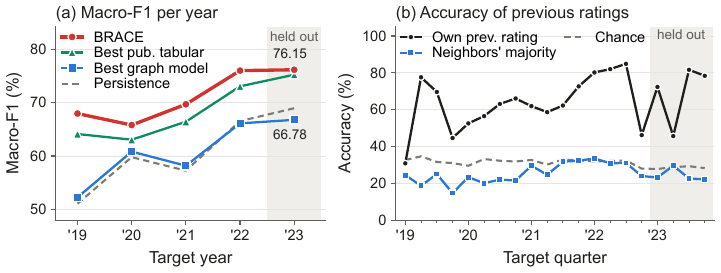}
\caption{\textbf{Without reading the estimated interbank network, BRACE beats every graph and tabular baseline across 2019--2023, including preregistered 2023.} (a)~Annual query Macro-F1 of BRACE, top tabular and graph baselines (Table~\ref{tab:main}), and persistence. (b)~Quarterly accuracy of own prior rating vs.\ neighbors' majority prior rating relative to chance $\sum_k p_k^2$.}
\label{fig:teaser}
\end{figure}

Within this landscape, our primary scope is \emph{partial-disclosure interbank credit rating}. Quarterly bank filings arrive staggered, leaving updated grades for early filers while most banks await assessment. On AI4Risk~\cite{interbank_repo}, Tang et~al.~\cite{tang2025dptap} formalize predicting four-class quarter-$t$ grades for $\approx 4{,}150$ query banks from lagged ratios and ratings of all 4,548 banks up to $t-1$ plus $\le 400$ disclosed support banks across 2016--2023 (Fig.~\ref{fig:teaser}a), alongside U.S.\ Call Reports~\cite{fdic_callreports} and observed anomaly graphs~\cite{tang2023gadbench}.

Prior methods span three classical paradigms: (1)~\emph{machine learning}, such as logistic regression and LightGBM~\cite{ke2017lightgbm,grinsztajn2022tree} fitted on historical panels or support banks; (2)~\emph{deep learning}, using MLPs and sequential encoders to model nonlinear attribute dynamics; and (3)~\emph{graph neural networks (GNNs)}, adapting spatial~\cite{kipf2017gcn,velickovic2018gat}, inductive~\cite{hamilton2017graphsage}, heterophily-aware~\cite{chien2021gprgnn}, and spectral/fraud GNNs~\cite{tang2022bwgnn,xiang2023gtan,chen2024consisgad} to minimum-density bank networks~\cite{anand2015filling} via TGAR~\cite{tgar}, PSAGNN~\cite{psagnn}, HFTCRNet~\cite{hftcrnet}, and DPTAP~\cite{tang2025dptap}.

Recently, \emph{Tabular Foundation Models (TFMs)} pretrained via in-context learning~\cite{muller2022pfn}---TabPFN~\cite{hollmann2025tabpfn,tabpfn25,tabpfn26}, TabICL~v2~\cite{tabiclv2}, and LimiX-16M~\cite{zhang2025limix}---have set a new state of the art under scarce labels, inferring posteriors in one pass to beat tuned GBDTs~\cite{baesens2026foundation}, while G2T-FM~\cite{eremeev2025g2tfm} and TFM4GAD~\cite{liu2026tfm4gad} adapt TFMs to graphs through feature augmentation and structural encodings, whereas GraphPFN~\cite{graphpfn} incorporates attention-based neighborhood aggregation and pretraining on synthetic graphs.

Despite these advances, existing GNN and TFM methods face sharp limitations. On interbank panels, confidential exposures force GNNs onto reconstructed graphs where neighbors' majority prior rating trails chance $\sum_k p_k^2$ in 18/20 quarters (Fig.~\ref{fig:teaser}b) under negative adjusted homophily~\cite{platonov2023characterizing}, leaving every GNN behind a matched MLP, whereas raw TFMs ignore \emph{cutoff drift} ($15\%$--$69\%$ re-cutting), \emph{label shift} ($\approx 17/36/34/12\%$), and non-decomposable Macro-F1~\cite{ye2012optimizing,narasimhan2024consistent}. On observed graphs, single-table TFM flattenings~\cite{eremeev2025g2tfm,liu2026tfm4gad} collapse multi-relation edges, seesaw across single backbones, alias disconnected component nullspaces, and lack noise gating.

To overcome these limitations, we propose \textbf{BRACE} (\emph{Bank Rating with Calibrated Ensembles}), decoupling multi-view posteriors from calibrated decisions (Fig.~\ref{fig:framework}). Stage~1 omits estimated bank edges to cross-fit nine posteriors pairing cutoff-realigned historical GBDT heads with history-context TabPFN~v2, while \textbf{BRACE-Ultra} pools TabPFN and LimiX-16M over ego, per-relation, and component-deflated Beta wavelet views on graphs. Stage~2 gates views via KKT-certified log-score pooling before BCTS~\cite{alexandari2020bcts}, panel MAP-EM~\cite{saerens2002adjusting,ye2024mapls}, and F1-aligned decoding.

Experiments across three banking panels and four anomaly graphs up to 5.78M nodes confirm both mechanisms. Support recalibration and MAP-EM alone lift tuned LightGBM from $56.37$ to $74.07$ ($\ge +7.29$ over all GNNs), while frozen BRACE hits $76.15$ on 2023---beating 12 published baselines, six inductive/hetero/spectral GNN arms ($66.78$ HFTCRNet, $65.90$ BWGNN), and Stage-2 TabPFN ($75.58$) across 2019--2023 and eight FDIC years. Meanwhile, KKT pooling assigns $5.6\%$ graph weight on bank edges ($70.45$ dev MF1) versus $27.9\%$--$100.0\%$ on anomaly graphs, topping TFM4GAD and SignGAD~\cite{huang2026signgad} ($94.70/80.69$ vs.\ $92.03/73.61$ on S; $92.34/75.06$ vs.\ $89.90/72.20$ unscreened \texttt{12c} on B) and halving $100\%$-rewiring loss.

Our main contributions are summarized as follows:
\begin{itemize}
\item \textbf{Diagnosing and solving partial-disclosure bank rating.} We show non-graph shifts dominate reconstructed bank graphs and design BRACE, achieving preregistered superiority on held-out 2023 (over 87 Base and 120 Base $\cup$ Matched arms) and transferring across eight U.S.\ Call Report years.
\item \textbf{Adaptive multi-view spectral-relational pooling.} We introduce \emph{BRACE-Ultra} with per-relation and component-deflated Beta wavelet views gated by KKT pooling, adapting automatically from noisy bank edges ($\le 0.37$ pt GNN rewiring sensitivity) to 5.78M-node anomaly graphs.
\item \textbf{Protocol forensics and reproducible audit.} We uncover evaluation confounds in bank GNN scripts, TFM4GAD eigensolvers, and cross-protocol GAD masks, releasing full code and KKT certificates.
\end{itemize}

%% file: sections/02_problem.tex
\section{Problem Formulation}
\label{sec:problem}

A panel tracks $N$ banks $\mathcal{V} = \{1, \dots, N\}$ over quarters $t = 1, \dots, T$ (Table~\ref{tab:data}), with ratios $\mathbf{x}_{i,t} \in \mathbb{R}^d$ ($d{=}70$), ordinal grade $y_{i,t} \in \{1, \dots, K\}$ ($K{=}4$; agency-derived, ML-imputed, and quarterly re-ranked in AI4Risk~\cite{interbank_repo}), and minimum-density network $\mathcal{G}_t = (\mathcal{V}, \mathcal{E}_t)$~\cite{anand2015filling}. Under DPTAP's protocol~\cite{tang2025dptap}, quarter $t$ reveals balanced support $\mathcal{S}_t \subset \mathcal{V}$ of $n_t = \min(100, m_t)$ banks/class from a random half of $\mathcal{V}$ ($|\mathcal{S}_t|{=}400$ from 2019Q4 onward; $292$--$372$ in 2019Q1--Q3), leaving $\mathcal{Q}_t = \mathcal{V} \setminus \mathcal{S}_t$ undisclosed. Predictors observe:
\begin{equation}
\mathcal{I}_t = \big\{\mathbf{x}_{i,s},\, y_{i,s} : i \in \mathcal{V},\, s \le t-1\big\} \cup \big\{y_{i,t} : i \in \mathcal{S}_t\big\},
\label{eq:info}
\end{equation}
plus $\mathcal{G}_{t-1}$ when graph views are enabled (omitted by baseline BRACE). Predictions on $\mathcal{Q}_t$ are scored by Macro-F1 ($\mathrm{MF1} = \allowbreak \frac{1}{K}\sum_{k=1}^K \allowbreak \frac{2\mathrm{TP}_k}{2\mathrm{TP}_k + \mathrm{FP}_k + \mathrm{FN}_k}$) and accuracy over 36 episodes/year (4 quarters $\times$ 3 support $\times$ 3 model seeds), forming nine quarterly-averaged seed-pair \emph{cells}. Three obstacles shape BRACE: \emph{cutoff drift} ($15\%$--$69\%$ transitions), \emph{scarce shifted disclosures} ($\le 400$ vs.\ $\approx 4{,}150$ skewed queries; zero shift on FDIC quartiles), and \emph{non-decomposable Macro-F1}~\cite{narasimhan2024consistent}.

\begin{table}[tb]
\centering
\caption{\textbf{Banking panels span skewed agency-derived grades and uniform quartiles.} Interbank and external U.S.\ Call Report summary.}
\label{tab:data}
\input{tables/tab_data}
\end{table}

%% file: tables/tab_data.tex
\scriptsize
\setlength{\tabcolsep}{5pt}
\begin{tabular}{@{}lccc@{}}
\toprule
Property & Interbank & FDIC-1 & FDIC-2 \\
\midrule
Source & AI4Risk & U.S.\ Call Reports & U.S.\ Call Reports \\
Banks / features & 4,548 / 70 & 5,657 / 70 & 4,887 / 70 \\
Years (dev / test) & 2016--23 (19--22 / 23) & 2008--14 (-- / 11--14) & 2012--18 (-- / 15--18) \\
Rating rule & \makecell[t]{Agency-derived, imputed,\\quarterly re-ranked} & Composite quartiles & Composite quartiles \\
Class shares (\%) & $\approx$17 / 36 / 34 / 12 & 25 / 25 / 25 / 25 & 25 / 25 / 25 / 25 \\
Quarterly change (\%) & 15--69 & 17--30 & 21--29 \\
Support / query & $\le$400 / $\approx$4,150 & 400 / 5,257 & 400 / 4,487 \\
Est.\ edges / quarter & $\approx$12.3k & 18.5k--27.8k & 21.1k--27.3k \\
\bottomrule
\end{tabular}

%% file: sections/03_method.tex
\section{Methodology: BRACE and BRACE-Ultra}
\label{sec:method}

\begin{figure}[t]
\centering
\includegraphics[width=\textwidth]{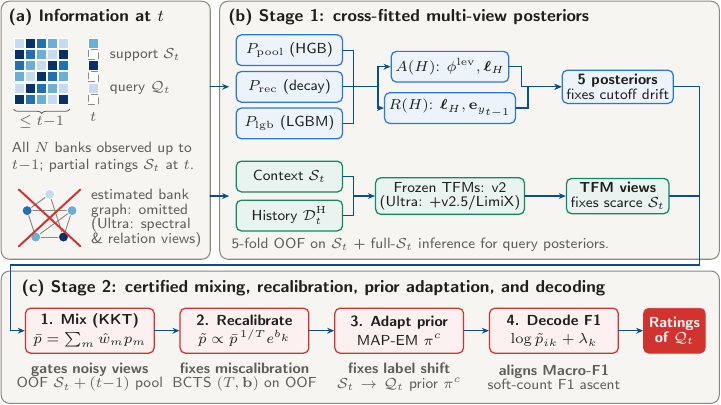}
\caption{\textbf{BRACE pairs obstacle-targeted posterior views with certified mixing and Macro-F1 calibration.} (a)~Information $\mathcal{I}_t$. (b)~Stage~1 fixes cutoff drift via realigned history heads and scarce $\mathcal{S}_t$ via frozen TFMs. (c)~Stage~2 gates noisy views via KKT pooling, recalibrates via BCTS, adapts priors via MAP-EM, and aligns decisions to Macro-F1.}
\label{fig:framework}
\end{figure}

Fitting one model on historical quarters imports stale cutoffs, fitting on $\mathcal{S}_t$ alone starves nonlinear learners and bakes in support priors, and end-to-end message passing overfits when edges are noisy. BRACE separates \emph{multi-view posterior generation} from \emph{certified pooling and F1 calibration} (Fig.~\ref{fig:framework}): Stage~1 maps $\mathcal{I}_t$ into $M$ cross-fitted posteriors ($M{=}9$ in BRACE; $M{=}15$--$18$ in Ultra-Bank; $M{=}10$--$12$ in Ultra on observed graphs), and Stage~2 gates views via KKT log-score pooling before BCTS, MAP-EM, and Macro-F1 decoding.

\subsection{Stage 1: Realigning History and Scarce Disclosures}
\label{sec:components}

\textbf{Support-realigned history heads (fixing cutoff drift).} Multi-year transitions $(\phi_i(\tau), y_{i,\tau})$ ($\tau \le t-1$) capture migration dynamics, yet raw historical posteriors misalign when quarter-$t$ cutoffs shift. Using inputs up to $t-1$, bank $i$ has lag vector $\phi^{\mathrm{lag}}_i(t) = [\mathbf{x}_{i,t-1}; \allowbreak \mathbf{e}_{y_{i,t-1}}; \allowbreak \mathbf{x}_{i,t-1}-\mathbf{x}_{i,t-2}; \allowbreak \mathbf{e}_{y_{i,t-2}}] \in \mathbb{R}^{148}$ and level vector $\phi^{\mathrm{lev}}_i(t) = [\mathrm{mm}_{t-1}(\mathbf{x}_{i,t-1}); \allowbreak \mathbf{e}_{y_{i,t-1}}] \in \mathbb{R}^{74}$, where $\mathbf{e}_k \in \{0,1\}^K$ is one-hot and $\mathrm{mm}_{t-1}$ min--max scales across $\mathcal{V}$ at $t-1$. Three class-balanced GBDTs fit historical transitions: $P_{\mathrm{pool}}$ (annual histogram GBDT on $\le 12$ quarters of $\phi^{\mathrm{lag}}$), $P_{\mathrm{rec}}$ (quarterly refit on 12 trailing quarters, half-life 4), and $P_{\mathrm{lgb}}$ (annual LightGBM~\cite{ke2017lightgbm} on 8 quarters of unscaled level rows, decay $0.8$). Five multinomial logistic heads on $\mathcal{S}_t$ realign $\boldsymbol{\ell}_H(i) = \log\max\{p_H(\cdot \mid i), 10^{-6}\}$ to current cutoffs: $A(P_{\mathrm{pool}}), A(P_{\mathrm{rec}})$ on $[\phi^{\mathrm{lev}}_i(t);\, \boldsymbol{\ell}_H(i)]$, and $R(P_{\mathrm{pool}}), R(P_{\mathrm{rec}}), R(P_{\mathrm{lgb}})$ on $[\boldsymbol{\ell}_H(i);\, \mathbf{e}_{y_{i,t-1}}]$.

\textbf{History-context foundation models (fixing scarce support).} Frozen TFMs (TabPFN~v2~\cite{hollmann2025tabpfn} by default, or TabPFN-2.6~\cite{tabpfn26} and LimiX-16M~\cite{zhang2025limix}) supply Bayesian priors on $\le 400$ rows: $\mathcal{S}_t$ alone yields \texttt{Lag} and \texttt{Level}, while \texttt{Lag+H} and \texttt{Level+H} prepend all $N$ prior transitions with a regime flag:
\begin{equation}
\mathcal{D}^{\mathrm{H}}_t = \big\{([\phi_j(t-1); 0],\, y_{j,t-1}) : j \in \mathcal{V}\big\} \cup \big\{([\phi_i(t); 1],\, y_{i,t}) : i \in \mathcal{S}_t\big\},
\label{eq:history}
\end{equation}
scoring queries at $[\phi_i(t); 1]$ ($\approx 4{,}950$ rows). One shared stratified 5-fold split of $\mathcal{S}_t$ yields OOF posteriors $p_m^{\mathrm{oof}}(\cdot \mid i)$ (keeping $t-1$ rows fixed) and full-$\mathcal{S}_t$ query posteriors $p_m(\cdot \mid i)$ (Table~\ref{tab:components}).

\begin{table}[tb]
\centering
\caption{\textbf{History-augmented TabPFN views draw 48\% of KKT weight in baseline BRACE.} Stage-1 components across 2019--2023 (180 episodes).}
\label{tab:components}
\input{tables/tab_components}
\end{table}

\subsection{Stage 2: View Gating, Calibration, and Macro-F1 Decoding}
\label{sec:stage2}

\textbf{KKT-certified simplex pool (gating noisy views).} Uniformly averaging views lets weak backbones or noisy graphs dilute strong predictors. BRACE forms $\bar{p}(k \mid i) = \allowbreak \sum_{m=1}^{M} \hat{w}_m p_m(k \mid i)$ on $\Delta^{M-1}$ by maximizing class-balanced log score~\cite{hall2007combining}:
\begin{equation}
\hat{\mathbf{w}} = \arg\max_{\mathbf{w} \in \Delta^{M-1}} \sum_{i \in \mathcal{R}} r_i \log \sum_{m=1}^{M} w_m\, p_m(y_i \mid i), \qquad \mathcal{R} = \mathcal{R}_S \cup \mathcal{R}_P,
\label{eq:weights}
\end{equation}
combining OOF support rows $\mathcal{R}_S$ with prior-quarter query posteriors $\mathcal{R}_P$ scored on $y_{i,t-1}$, each source class-balanced via $1/n_{s,y_i}$ ($s \in \{S, P\}$) and normalized to $1/2$ total mass (2019Q1 uses uniform $1/|\mathcal{S}_t|$ on $\mathcal{R}_S$ alone). Active-set Newton steps certify KKT optimality ($g_m = \allowbreak \sum_i r_i p_m(y_i \mid i) / \allowbreak \bar{p}(y_i \mid i) = 1$ if $\hat{w}_m > 0$ and $\le 1$ else, to $10^{-7}$; max residual $4.9 \times 10^{-9}$ BRACE, $2.2 \times 10^{-8}$ Ultra-Bank), pruning dominated views to exact $\hat{w}_m{=}0$.

\textbf{BCTS, MAP-EM, and soft-count F1 decoding.} Fitting $(T, \mathbf{b})$ ($b_1{=}0$) by maximum likelihood on $\bar{p}^{\mathrm{oof}}$ yields BCTS posteriors~\cite{alexandari2020bcts} $\tilde{p}(k \mid i) \propto \bar{p}(k \mid i)^{1/T} e^{b_k}$. Extending maximum-likelihood EM label-shift estimation~\cite{saerens2002adjusting} with a symmetric $\mathrm{Dir}(c{+}1,\dots,c{+}1)$ prior~\cite{ye2024mapls}, regularized MAP-EM with strength $c \in \{0, 10, 30, 100, 300, 1000, \infty\}$ adapts $\tilde{p}$ to $\mathcal{Q}_t$ from $\pi^{(0)}_k = 1/K$ ($c{=}0$ recovers MLE; $c{=}\infty$ preserves uniform priors when no shift exists):
\begin{align}
\gamma^{(j)}_{ik} &= \frac{\tilde{p}(k \mid i)\, \pi^{(j)}_k}{\sum_{k'} \tilde{p}(k' \mid i)\, \pi^{(j)}_{k'}}, \qquad
\pi^{(j+1)}_k = \frac{\sum_{i \in \mathcal{Q}_t} \gamma^{(j)}_{ik} + c}{|\mathcal{Q}_t| + Kc}, \label{eq:mapem}\\
\mathrm{EF1}(\hat{\mathbf{y}}; \boldsymbol{\rho}) &= \frac{1}{K}\sum_{k=1}^{K}\frac{2\sum_{i:\hat{y}_i=k}\rho_{ik}}{\sum_{i \in \mathcal{Q}_t}\rho_{ik} + |\{i:\hat{y}_i=k\}|}. \label{eq:ef1}
\end{align}
Denoting converged MAP-EM posteriors and priors at strength $c$ by $\boldsymbol{\gamma}(c)$ and $\boldsymbol{\pi}(c)$, BRACE sets $\boldsymbol{\rho} = \boldsymbol{\gamma}(10)$ to target Macro-F1~\cite{ye2012optimizing,lipton2014thresholding}, selects $c^\star$ maximizing $\mathrm{EF1}(\arg\max_k \gamma_{ik}(c); \boldsymbol{\rho})$, and from $\boldsymbol{\lambda} = \log(K\boldsymbol{\pi}(c^\star))$ runs coordinate ascent on $\hat{y}_i(\boldsymbol{\lambda}) = \arg\max_k(\log \tilde{p}(k \mid i) + \lambda_k)$ over step grid $\delta \in \{-1.50, -1.45, \dots, 1.50\}$ ($\le 30$ sweeps; 11 s/yr vs.\ 41--60 min Stage~1).

\subsection{Extending Stage 1 to Informative Graphs: BRACE-Ultra}
\label{sec:gad_method}

\textbf{Why naive graph flattenings break.} When observed graphs $\mathcal{G} = (\mathcal{V}, \{\mathcal{E}_r\}_{r=1}^R)$ carry fraud signal ($K{=}2$), 1-hop relation means (\emph{BRACE $\mathrm{C0}$}; 2-hop + OOF neighbor rates in Protocol~G $\mathrm{C4}$) smooth away heterophilous anomalies. Meanwhile, single-table TFM4GAD~\cite{liu2026tfm4gad} suffers from three bottlenecks: it collapses multi-relation edges into one graph, relies on one TFM backbone whose inductive bias seesaws across graph densities, and slices raw Laplacian eigensolvers without deflating disconnected components, so on a graph with $C_{\mathrm{comp}}$ components ($3{,}437$ on T-Social, $26$ on YelpChi), leading slices capture trivial degree-scaled component indicators ($\lambda{=}0$) rather than structural Fiedler modes.

\textbf{Multi-scale views and nullspace deflation.} \emph{BRACE-Ultra} builds four label-free views on $\tilde{L} = I - \tilde{D}^{-1/2}(A_{\mathrm{sym}}+I)\tilde{D}^{-1/2}$ (schemas in Appendix~\ref{app:schemas}): (i)~\texttt{ego} ($\mathbf{x}_i$), preserving pure attributes when edges are noisy; (ii)~\texttt{rel-lowpass}, recording per-relation 1-hop moments, residuals $\mathbf{x}_i - \boldsymbol{\mu}_{i,r}$, and 1/2-hop SAGE means; (iii)~\texttt{spec-wavelet}, combining $C{=}3$ Beta band-pass wavelets~\cite{tang2022bwgnn} (plus per-relation high-pass filters when $R{>}1$), random-walk structural stats, and \emph{component-deflated} \texttt{LapPE}($k{=}16$), projecting out the $C_{\mathrm{comp}}$ component zero modes ($M_\perp = M - 2U_0 U_0^\top$) before Lanczos to recover the 16 lowest nonzero Laplacian eigenvectors (sign-fixed); and (iv)~\texttt{joint-fusion}, combining residuals, 2-hop means, wavelets, and \texttt{LapPE}.

\textbf{Cross-foundation pooling and bank exposure views.} Because $\ell_2$ logistic heads over-smooth OOF log loss under $<100$ labels (\S\ref{sec:rq6}), BRACE-Ultra pools pure TFMs across TabPFN~v2, v2.5, and LimiX-16M ($M{=}12$ symmetric $3{\times}4$ pool on B; $M{=}10$ S-val pool dropping v2.5/LimiX \texttt{ego} on S) via Eq.~\eqref{eq:weights} and box-constrained BCTS ($T \in [0.1, e^{10}]$) under \emph{Train-Only} (\textbf{TO}, \texttt{Val=0} at fit: train-OOF weights, BCTS, threshold) and \emph{Train+Val} (\textbf{TV}, $1/2$ train-OOF + $1/2$ val weights, val threshold). On 2016--2022 bank panels (2023/FDIC sealed), \emph{BRACE-Ultra-Bank} pools counterparty momentum and $\log(1+|W|)$ Beta wavelets (\texttt{net-sl} $150\mathrm{d}$, \texttt{net-hssl} $135\mathrm{d}$; Appendix~\ref{app:schemas}).

%% file: tables/tab_components.tex
\scriptsize
\setlength{\tabcolsep}{4.5pt}
\begin{tabular}{@{}lllcc@{}}
\toprule
Component & Input features (dim.) & Context & Mean wt. & Active (\%) \\
\midrule
$A(P_{\mathrm{pool}})$ & $\phi^{\mathrm{lev}}$, $\boldsymbol{\ell}$ (78) & $\mathcal{S}_t$ & 0.042 & 38 \\
$A(P_{\mathrm{rec}})$ & $\phi^{\mathrm{lev}}$, $\boldsymbol{\ell}$ (78) & $\mathcal{S}_t$ & 0.083 & 57 \\
$R(P_{\mathrm{pool}})$ & $\boldsymbol{\ell}$, $\mathbf{e}_{y_{t-1}}$ (8) & $\mathcal{S}_t$ & 0.003 & 11 \\
$R(P_{\mathrm{rec}})$ & $\boldsymbol{\ell}$, $\mathbf{e}_{y_{t-1}}$ (8) & $\mathcal{S}_t$ & 0.013 & 14 \\
$R(P_{\mathrm{lgb}})$ & $\boldsymbol{\ell}$, $\mathbf{e}_{y_{t-1}}$ (8) & $\mathcal{S}_t$ & 0.044 & 54 \\
\addlinespace[1.5pt]
TabPFN-Lag & $\phi^{\mathrm{lag}}$ (148) & $\mathcal{S}_t$ & 0.227 & 81 \\
TabPFN-Level & $\phi^{\mathrm{lev}}$ (74) & $\mathcal{S}_t$ & 0.110 & 51 \\
TabPFN-Lag+H & $\phi^{\mathrm{lag}}$, flag (149) & $\mathcal{D}^{\mathrm{H}}_t$ & 0.329 & 85 \\
TabPFN-Level+H & $\phi^{\mathrm{lev}}$, flag (75) & $\mathcal{D}^{\mathrm{H}}_t$ & 0.148 & 61 \\
\bottomrule
\end{tabular}

%% file: sections/04_setup.tex
\section{Experimental Setup}
\label{sec:setup}

Our evaluation has two parts: \textbf{Part~I (Interbank Diagnosis \& Solution)} tests 2023 and Stage-2 controls (\textbf{RQ1}), zero-shift FDIC transfer (\textbf{RQ2}), ablations (\textbf{RQ3}), and GNN falsification (\textbf{RQ4}); \textbf{Part~II (Counterfactual Graph Scaling)} tests KKT gating on bank (\textbf{RQ5}) and anomaly graphs (\textbf{RQ6}).

\textbf{Benchmarks and label budgets.} Table~\ref{tab:data} summarizes \emph{Interbank}~\cite{interbank_repo} (2019--2022 dev, 2023 test; 2016--2018 history) and balanced U.S.\ FDIC Call Reports~\cite{fdic_callreports} \emph{FDIC-1} (2011--2014) and \emph{FDIC-2} (2015--2018), graded by quartiles (\texttt{25/25/25/25\%}, zero shift) with minimum-density edges~\cite{anand2015filling}. All methods run 36 episodes/yr under $\mathcal{I}_t$, scored on $\mathcal{Q}_t$ (upstream TGAR/HFTCRNet score $\mathcal{S}_t \cup \mathcal{Q}_t$ at the test-best epoch; $\mathcal{Q}_t$-only readouts put $22/24$ non-test readings below published values). For \textbf{RQ6}, we test four GADBench graphs~\cite{tang2023gadbench} (\emph{Amazon}: $11.9\mathrm{k}$, $R{=}3$, $C_{\mathrm{comp}}{=}1$; \emph{YelpChi}: $46.0\mathrm{k}$, $R{=}3$, $C_{\mathrm{comp}}{=}26$; \emph{T-Finance}: $39.4\mathrm{k}$, $R{=}1$, $C_{\mathrm{comp}}{=}4$; \emph{T-Social}: $5.78\mathrm{M}$ nodes, $146.2\mathrm{M}$ edges, $R{=}1$, $C_{\mathrm{comp}}{=}3{,}437$): \emph{Protocol~B}~\cite{tang2023gadbench,liu2026tfm4gad} (10 splits, cols $10$--$19$: $100$ train w/ $20$ anomalies; \texttt{Val=0} vs.\ \texttt{Val=100}), \emph{Protocol~S}~\cite{chen2024consisgad} (5 seeds: $86/459/393/578$ train; \texttt{Val=0} vs.\ $33\%$ val [$2.8\mathrm{k}/15.0\mathrm{k}/12.9\mathrm{k}/1.91\mathrm{M}$], $0$ S-test overlap), and \emph{Protocol~G}~\cite{huang2026signgad} (10 seeds, $1/49.5/49.5\%$).

\textbf{Matched baselines and protocol hygiene.} Preregistered \emph{Published} (12 arms) covers TGAR~\cite{tgar}, HFTCRNet~\cite{hftcrnet}, GCN~\cite{kipf2017gcn}, and GAT~\cite{velickovic2018gat} (final and $\mathcal{S}_t$-best, 8 arms), default/tuned LightGBM~\cite{ke2017lightgbm} (2), and lag/level TabPFN~v2~\cite{hollmann2025tabpfn} (2), plus Persistence and Support LR (Table~\ref{tab:main}). Before FDIC-1/2023, we froze \emph{Base} (87 single-family arms) and \emph{Matched} (33 arms pairing single, history, and TabPFN arms with BRACE's F1 decoder, $\pm$BCTS). For cellwise lead $d$, we preregistered operational criteria chosen for this benchmark: \textbf{superior} if $\bar{d} > 0$ ($d > 0$ in $\ge 7/9$ cells) and \textbf{non-inferior} if $\bar{d} \ge -0.5$ ($d \ge -0.5$ in $\ge 7/9$ cells). Reference~\cite{piaggio2012noninferiority} provides general guidance on prespecifying and reporting non-inferiority margins. Post-2023 arms (\textdagger) add TabPFN-2.6~\cite{tabpfn26} (registered on FDIC-2), LimiX-16M~\cite{zhang2025limix}, six inductive/hetero/spectral GNN arms (inductive GraphSAGE~\cite{hamilton2017graphsage}, spectral $d{=}2$ BWGNN~\cite{tang2022bwgnn}, heterophily-aware $K{=}10$ GPR-GNN~\cite{chien2021gprgnn} on $\mathcal{G}_{t-2}\to\mathcal{G}_{t-1}$; 2019Q1 GCN parity $<10^{-6}$, gap $\le 0.67$ pts), and a 300-epoch transductive $\phi^{\mathrm{lev}}$ test on $\mathcal{G}_{t-1}$ vs.\ a 256-unit MLP; \emph{BRACE-Ultra-Bank} (\textdaggerdbl) blocked $\ge$2023Q1/FDIC. On GAD, we test TFM4GAD~\cite{liu2026tfm4gad} ($0.00$--$0.13$ diff across 28 Table~2 means), GADBench max~\cite{tang2023gadbench}, MLP, GCN, GraphSAGE, GATv2~\cite{brody2022gatv2}, GTAN~\cite{xiang2023gtan}, DGA-GNN~\cite{duan2024dgagnn}, DiG-In-GNN~\cite{zhang2024digingnn}, ConsisGAD~\cite{chen2024consisgad}, SpaceGNN~\cite{dong2025spacegnn}, and SignGAD~\cite{huang2026signgad} (\texttt{refit\_off} on B/S). Since S-val overlaps B-test ($199{,}990$--$199{,}998/200\mathrm{k}$ T-Social S-val nodes), Table~\ref{tab:gad}a uses unscreened $3{\times}4$ \texttt{12c} (no S-val selection) on B (\textsuperscript{\S}$\mathrm{C0}$ unranked); on S (\textsuperscript{*}), \texttt{Val=0} uses zero val labels at fit on the S-val 10-TFM pool.

%% file: sections/05_results.tex
\section{Experimental Results}
\label{sec:results}

\subsection{RQ1: Decomposing Gains Over Graph and Tabular Baselines}
\label{sec:rq1}

\begin{table}[t]
\centering
\caption{\textbf{Output alignment rescues tabular baselines over GNNs, and Stage-1 history pooling lifts BRACE past Stage-2-calibrated TabPFN.} Query Macro-F1 and Acc ($\times 100$, 36 episodes/yr); bottom: cellwise MF1 leads $d$. \textsuperscript{\textdagger}Post-2023. \textsuperscript{\textdaggerdbl}2019--22 (2023 sealed). \texttt{R2}: support head $R(\cdot)$; \texttt{D*}: dev MAP-EM ($c^\star{=}300$), argmax.}
\label{tab:main}
\input{tables/tab_main}
\end{table}

\begin{figure}[t]
\centering
\includegraphics[width=\textwidth]{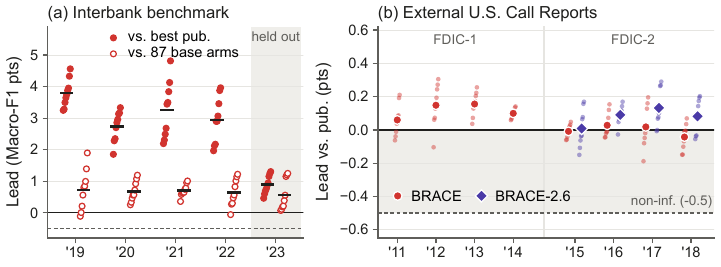}
\caption{\textbf{BRACE wins every Interbank cell against published baselines and stays superior or non-inferior across all eight external Call Report years.} (a)~Interbank cell leads over Published (filled) and Base (open). (b)~Post-hoc external FDIC-1 and FDIC-2 leads over Published relative to the $-0.5$ non-inferiority margin.}
\label{fig:leads}
\end{figure}

\textbf{Three-step decomposition of the gain.} Table~\ref{tab:main} isolates three effects. \emph{First, tabular priors beat every GNN}: adding inductive GraphSAGE ($56.78$ dev MF1), spectral BWGNN ($59.23$ dev, $+12.30$ over GCN and ahead of TGAR and HFTCRNet), and heterophily-aware GPR-GNN ($28.07$ under 2000-epoch decay) changes $0/45$ cell maxima over 2019--2023, leaving raw TabPFN~v2 ($66.67$ dev / $75.23$ in 2023) on top. \emph{Second, output alignment rescues single learners}: tuned LightGBM with support recalibration $R(\cdot)$ (\texttt{R2}) and dev MAP-EM ($c^\star{=}300$, argmax; \texttt{D*}) hits $66.41$ dev MF1 ($+10.75$ vs.\ argmax) and $74.07$ on 2023 ($+17.70$), topping every GNN by $\ge 7.29$ pts without F1 decoding or a TFM; Stage~2 on standalone TabPFN~v2 lifts MF1 from $66.67 \to 68.67$ dev ($+2.00$) and $75.23 \to 75.58$ on 2023. \emph{Third, Stage-1 history and KKT pooling top calibrated TabPFN}: frozen BRACE hits $69.86$ dev MF1 and $76.15$ ($78.36\%$ Acc) on held-out 2023, beating Stage-2 \texttt{TabPFN-lag} by $+1.18$ dev pts ($35/36$ cells) and $+0.56$ on 2023 ($9/9$ cells, superior yearly), and BWGNN by $+10.63$ dev pts ($180/180$ episodes; $+5.71$/$+5.65$ over BWGNN+\texttt{R2}/GraphSAGE+\texttt{R2}). Table~\ref{tab:heldout} and Fig.~\ref{fig:leads}a confirm preregistered 2023 superiority over Base ($87$ arms, $+0.57$, $9/9$; 95\% CI $[+0.30, +0.68]$), Base $\cup$ Matched ($120$ arms, $+0.40$, $9/9$), Published ($+0.89$), and the 18-arm set ($+0.89$), lifting minority tiers~3--4 by $+8.2$/$+8.0$ F1 and transition precision ($57.8\%$--$66.2\%$ vs.\ $49.8\%$--$55.6\%$).

\begin{table}[tb]
\centering
\caption{\textbf{Frozen BRACE achieves preregistered superiority on held-out 2023, including over Matched Stage-2 TabPFN arms, and retains it against the 18-arm GNN set (\textsuperscript{\textdagger})} ($d$ in MF1 pts over 9 cells).}
\label{tab:heldout}
\input{tables/tab_heldout}
\end{table}

\subsection{RQ2: Transfer Under Low Churn and Zero Prior Shift}
\label{sec:rq2}

\begin{table}[tb]
\centering
\caption{\textbf{BRACE and BRACE-2.6 stay superior or non-inferior across all eight zero-prior-shift U.S.\ Call Report years.} Top: query Macro-F1 ($\times 100$). Bottom: mean cellwise lead $d$ and verdict (\textsuperscript{S}~superior, \textsuperscript{N}~non-inferior; vs.\ Base on FDIC-1 and BRACE-2.6 vs.\ BRACE on FDIC-2 are preregistered, vs.\ pub.\ post-hoc).}
\label{tab:fdic}
\input{tables/tab_fdic}
\end{table}

\textbf{Robustness without synthetic support-to-query shift.} FDIC-1 and FDIC-2 test whether BRACE relies on AI4Risk's balanced-support shift (Table~\ref{tab:fdic}, Fig.~\ref{fig:leads}b): grades are exact quartiles (\texttt{25/25/25/25\%}, zero shift), while lower churn ($17\%$--$30\%$) lifts Persistence to $75.60$--$80.52$ MF1 and narrows headroom. Without retuning, MAP-EM shrinks toward uniform priors, keeping BRACE preregistered non-inferior to all 87 Base arms on FDIC-1 (95\% CI $\ge -0.18$, all $36/36$ cells $\ge -0.23$ pts; superior in 2012, $7/9$ cells) and superior to Published in 2012--2014; on FDIC-2, BRACE-2.6 is preregistered superior to BRACE in 2016--2018 and superior to Published in 2016--2017 ($8/9$, $7/9$ cells; all $36/36$ cells $\ge -0.16$), while GNNs trail by $0.9$--$21.2$ pts.

\subsection{RQ3: Stage-2 Ablation on Stored Posteriors}
\label{sec:rq3}

\begin{figure}[t]
\centering
\includegraphics[width=\textwidth]{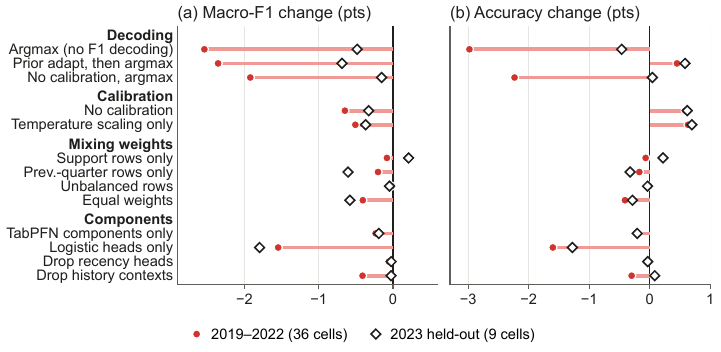}
\caption{\textbf{Soft-count Macro-F1 decoding drives the largest Stage-2 gain, complemented by BCTS calibration and KKT pooling.} Change in (a)~Macro-F1 and (b)~accuracy on stored posteriors (2023 diagnostics are post-hoc).}
\label{fig:ablation}
\end{figure}

\textbf{F1 decoding outweighs prior shift alone.} Fixing Stage-1 posteriors (Fig.~\ref{fig:ablation}), reverting from soft-count F1 ascent to calibrated argmax drops MF1 by $2.54$ dev pts ($36/36$ cells) and $0.48$ on 2023 ($9/9$); MAP-EM + argmax still loses $2.35$/$0.69$, showing F1 coordinate ascent (rather than prior adjustment alone) drives most of Stage~2. Dropping BCTS or using scalar $T$ costs $0.51$--$0.65$ dev and $0.33$--$0.37$ in 2023 (lifting Acc by $0.59$--$0.70$) and worsens prior-adjusted NLL ($0.592$--$0.595$ vs.\ $0.579$ dev; $0.558$--$0.560$ vs.\ $0.550$ in 2023), Brier, and ECE. KKT pooling tops the best single arm (\texttt{TabPFN-Level+H} by $+0.90$ dev; \texttt{TabPFN-Lag} by $+0.56$ in 2023), while equal weights lose $0.41$/$0.58$, logistic heads alone lose $1.55$/$1.80$, and dropping history costs $0.41$ dev.

\subsection{RQ4: Why Message Passing Fails on Reconstructed Bank Graphs}
\label{sec:rq4}

\begin{figure}[t]
\centering
\includegraphics[width=\textwidth]{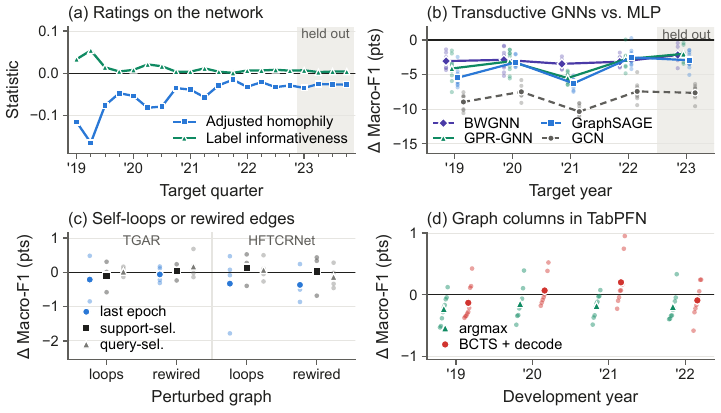}
\caption{\textbf{Controlled falsification shows negative or negligible rating signal across homophilous, inductive, heterophily-aware, and spectral GNNs on $\mathcal{G}_{t-1}$.} (a)~Adjusted homophily and label informativeness. (b)~Transductive $\phi^{\mathrm{lev}}$ GNNs minus a 256-unit MLP. (c)~Self-loops or rewired edges. (d)~Lag-row TabPFN with 154 graph cols.}
\label{fig:graph}
\end{figure}

\textbf{Five falsifications rule out homophily mismatch.} On $\mathcal{G}_{t-1}$ (Fig.~\ref{fig:graph}): (i)~adjusted homophily~\cite{platonov2023characterizing} is negative in all 20 quarters ($-0.165$ to $-0.015$) with label informativeness $0.001$--$0.054$ (Fig.~\ref{fig:graph}a); (ii)~in transductive tests (74-dim $\phi^{\mathrm{lev}}$, $80/20$ $\mathcal{S}_t$ split, 300 balanced epochs), GraphSAGE ($60.32$ dev / $69.97$ in 2023), GPR-GNN ($60.82$ / $70.89$), and BWGNN ($61.57$ / $70.73$) top GCN ($56.14$ / $65.27$), yet \emph{all four trail a 256-unit MLP yearly} ($64.70$ / $72.91$; $-3.14$ to $-8.56$ dev MF1, $\le 1/36$ positive dev cells; Fig.~\ref{fig:graph}b), while GPR-GNN drives odd-hop sums negative in $144/144$ dev episodes (mean $-0.594$) on 70 ratios and, given $\mathbf{e}_{y_{t-1}}$, lifts ego $\gamma_0$ ($0.100 \to +0.377$) while canceling $\sum_{k=1}^{10}\gamma_k$ ($0.900 \to 0.128$); (iii)~Correct and Smooth~\cite{huang2021cs} loses $0.99$/$7.68$ pts at $\alpha_s{=}0.2$/$0.5$; (iv)~loops/rewirings in TGAR/HFTCRNet move dev MF1 by $\le 0.37$ pts (Fig.~\ref{fig:graph}c); and (v)~154 neighbor cols in TabPFN shift MF1 by $-0.19$/$+0.01$ (Fig.~\ref{fig:graph}d).

\subsection{RQ5: Counterfactual Part A: Bank Exposure Wavelets}
\label{sec:rq5}

\begin{figure}[t]
\centering
\includegraphics[width=\textwidth]{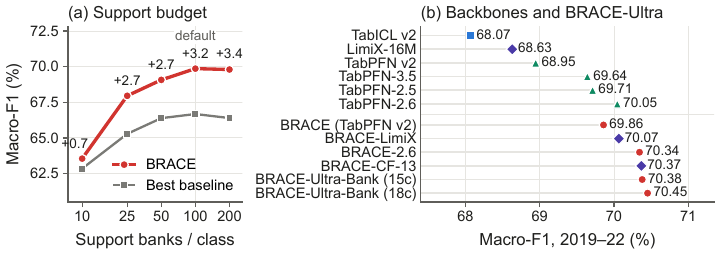}
\caption{\textbf{BRACE scales from 25 banks/class; KKT gives 5.6\%--8.2\% mean weight to wavelets.} (a)~Support sweep (2019--22). (b)~Backbones and pools.}
\label{fig:backbones}
\end{figure}

\textbf{KKT gating down-weights weak exposure signal without hurting tabular priors.} For $k \in \{10, 25, 50, 100, 200\}$/class (Fig.~\ref{fig:backbones}a), BRACE wins $141/144$ cells from $k{=}25$ up ($+2.67$ at $k{=}25$, $+3.41$ at $k{=}200$). With \texttt{Lag+H} + Stage~2 (Fig.~\ref{fig:backbones}b), backbones hit $68.07$ dev MF1 for TabICL~v2~\cite{tabiclv2}, $68.63$ LimiX-16M~\cite{zhang2025limix} ($69.11$ \texttt{Level+H}), $68.95$ TabPFN~v2, $69.64$ v3.5~\cite{tabpfn35}, $69.71$ v2.5~\cite{tabpfn25}, and $70.05$ v2.6~\cite{tabpfn26}; stacking hits $70.07$ in \emph{BRACE-LimiX} ($76.15$ on 2023), $70.34$ in \emph{BRACE-2.6} ($76.79$ on 2023), and $70.37$ in \emph{BRACE-CF-13} ($20.4\%$ LimiX wt.; $70.43$ +v3.5). Adding counterparty momentum and $\log(1+|W|)$ wavelets (\emph{BRACE-Ultra-Bank}, 2023 sealed), \texttt{net-sl} ($68.62$) and \texttt{net-hssl} ($69.22$) edge graph-free peers ($+0.37$/$+0.11$), while KKT assigns $8.2\%$ ($15\mathrm{c}$, $70.38$, $+0.01$ vs.\ $13\mathrm{c}$; $70.39$ w/ net head) and $5.6\%$ ($18\mathrm{c}$, $70.45$ MF1, $75.02\%$ Acc; $+0.02$ vs.\ $16\mathrm{c}$, $+0.59$ vs.\ BRACE, $115/144$) mean weight.

\subsection{RQ6: Counterfactual Part B: Informative Anomaly Graphs}
\label{sec:rq6}

\begin{table}[!t]
\centering
\caption{\textbf{On observed graphs, KKT pooling assigns 27.9\%--100.0\% weight to spectral/relational views and leads TFM4GAD in AUROC/AUPRC.} Test AUROC/AUPRC (\%; stds and all 25 GADBench / 18 ConsisGAD baselines in Appendix~\ref{app:gad}). \texttt{12c}: unscreened $3{\times}4$ TFM pool; \textsuperscript{*}10-TFM (S), \textsuperscript{\S}$\mathrm{C0}$ (unranked on B): S-val pools; SignGAD: \texttt{refit\_off}.}
\label{tab:gad}
\input{tables/tab_gad}
\end{table}

\begin{figure}[!t]
\centering
\includegraphics[width=\textwidth]{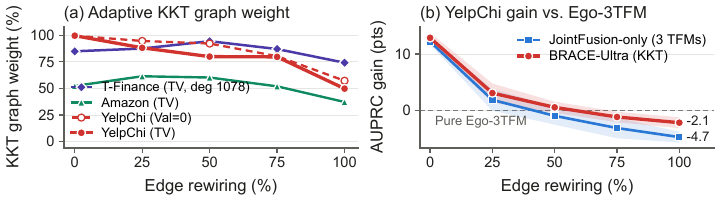}
\caption{\textbf{Edge rewiring lowers KKT graph weight and halves YelpChi noise loss vs.\ \texttt{Ego-3TFM}.} Splits $0$--$4$: (a)~KKT weight; (b)~YelpChi gain ($\pm 1$ SEM).}
\label{fig:rewire}
\end{figure}

\textbf{Spectral forensics of TFM4GAD.} On skewed anomaly graphs ($3\%$--$14.5\%$ anomalies, Table~\ref{tab:gad}), TFM4GAD~\cite{liu2026tfm4gad} (\texttt{Cloudy1225/TFM4GAD}) hits $\approx 97/87\%$ B/T-Social AUROC/AUPRC yet has three flaws: (i)~relation collapse caps YelpChi at $70.52$--$71.65 / 30.74$--$31.85$ (B) and $81.06$--$81.32 / 41.58$--$44.40$ (S); (ii)~backbones seesaw ($91.81/62.43$ LimiX S/Amazon; $94.46/79.52$ v2.5 B/T-Finance); and (iii)~on 3,437-component T-Social, \texttt{4bb8a51} runs \texttt{randomized\_\allowbreak svd}($L_{\mathrm{sym}}$) on \emph{highest-frequency} modes ($\sigma \in [1.80, 1.82]$), while \texttt{PR~\#1} (\texttt{5bb1367}) slices \texttt{eigsh}($2I - L_{\mathrm{sym}}, k{=}17, \texttt{tol}{=}10^{-4}$) without deflation, aliasing zero/$\sim 10^{-5}$ modes ($0.30$ drift; Split-0 LimiX-16M $84.16 \to 81.93$, 2M $71.88$ vs.\ $83.26$ PRC).

\textbf{Disjoint Protocol~S and unscreened Protocol~B.} On Protocol~S (\textsuperscript{*}10-TFM S-val pool), \emph{BRACE-Ultra (TO)} (\texttt{Val=0} fit) hits $94.70/80.69$ AUROC/AUPRC, beating TFM4GAD-v2.5 ($+2.67/+7.08$) and LimiX ($+3.07/+10.32$) as YelpChi hits $90.35/65.24$ ($+20.84$ PRC); at \texttt{Val=33\%}, it hits $94.99/82.36$ ($88.24$ MF1; T-Social $94.16/50.15 \to 97.29/90.93$, vs.\ GNNs $34.46$--$65.81$ PRC). On Protocol~B (\emph{no S-val selection}), $3{\times}4$ \emph{BRACE-Ultra (12c, TO)} (\texttt{Val=0}) hits $92.34/75.06$ ($80.23$ MF1; $+2.44/+2.86$ vs.\ LimiX, $+2.26$ vs.\ $72.80$ oracle PRC) while \texttt{Val=100} hits $92.65/76.21$, matching S-val 10-TFM ($92.38/75.00$, $92.66/76.22$); \texttt{16c} ($92.20/75.01$, $92.62/76.28$) and B-val nested selection ($91.00/\allowbreak 72.29$ vs.\ $87.54/67.27$ LimiX, 3 graphs) win. YelpChi collapse (10-TFM TV) costs $-11.31$/$-7.70$ (B/S) PRC; dropping LR adds $+13.12$ PRC on S/Amazon TO ($83.89$ vs.\ $70.77$); on G ($n{=}10$), BRACE ($\mathrm{C4}$) tops SignGAD~\cite{huang2026signgad} ($98.98/92.60$, $97.38/90.30$, $98.16/93.34$, $99.72/96.38$ ROC/MF1).

\textbf{Controlled edge rewiring.} Rewiring $\rho \in \{0,\allowbreak 25,\allowbreak 50,\allowbreak 75,\allowbreak 100\}\%$ of edges to Erd\H{o}s--R\'{e}nyi graphs (fixed $\mathbf{X}$, splits $0$--$4$, \texttt{12c}; Fig.~\ref{fig:rewire}), YelpChi ($R{=}3$) KKT graph weight falls monotonically under TV ($99.58 \to 88.19 \to 79.91 \to 79.67 \to 49.83\%$, \texttt{ego} $0.42\% \to 50.17\%$) and \texttt{Val=0} ($99.52 \to 94.63 \to 92.18 \to 80.32 \to 57.24\%$); at $100\%$, \texttt{12c} halves \texttt{JointFusion-\allowbreak only} loss vs.\ \texttt{Ego-3TFM} ($-2.13$ vs.\ $-4.67$ PRC; $-0.74$ vs.\ $-2.46$ TV, $-0.12$ vs.\ $-1.64$ \texttt{Val=0} MF1). T-Finance (deg.\ $1{,}078$, $\approx 270$ neighbors at $75\%$) keeps $+7.11$ PRC ($84.05$ vs.\ $76.94$) and $84.91\%$--$94.41\%$ weight ($74.18\%$ at $100\%$, $-2.23$ vs.\ $-4.06$); Amazon (\texttt{Ego-3TFM} $87.28$) drops $52.34\% \to 37.27\%$ ($\Delta\mathrm{PRC} \in [-0.12, +0.69]$).

%% file: tables/tab_main.tex
\scriptsize
\setlength{\tabcolsep}{2.9pt}
\begin{tabular}{@{}lcccccccc@{}}
\toprule
 & \multicolumn{4}{c}{Annual dev MF1} & \multicolumn{2}{c}{2019--22 Mean} & \multicolumn{2}{c}{2023 (held out)} \\
\cmidrule(lr){2-5}\cmidrule(lr){6-7}\cmidrule(lr){8-9}
Method & 2019 & 2020 & 2021 & 2022 & MF1 & Acc & MF1 & Acc \\
\midrule
\multicolumn{9}{@{}l}{\emph{Graph models (estimated $\mathcal{G}_{t-2}\to\mathcal{G}_{t-1}$)}} \\
TGAR~\cite{tgar} & 52.28 & 60.51 & 57.55 & 64.04 & 58.59 & 67.66 & 63.81 & 68.49 \\
HFTCRNet~\cite{hftcrnet} & 51.47 & 60.39 & 57.11 & 64.64 & 58.40 & 63.93 & 66.78 & 68.01 \\
GCN~\cite{kipf2017gcn} & 40.11 & 47.53 & 48.70 & 51.40 & 46.93 & 60.65 & 52.22 & 58.90 \\
GAT~\cite{velickovic2018gat} & 44.05 & 54.75 & 53.04 & 56.21 & 52.01 & 63.77 & 57.23 & 63.29 \\
BWGNN\textsuperscript{\textdagger}~\cite{tang2022bwgnn} & 51.78 & 60.84 & 58.20 & 66.12 & 59.23 & 66.82 & 65.90 & 68.54 \\
GraphSAGE\textsuperscript{\textdagger}~\cite{hamilton2017graphsage} & 50.00 & 58.76 & 57.18 & 61.17 & 56.78 & 67.31 & 62.02 & 67.35 \\
GPR-GNN\textsuperscript{\textdagger}~\cite{chien2021gprgnn} & 32.61 & 31.15 & 27.05 & 21.47 & 28.07 & 48.22 & 30.54 & 46.24 \\
\addlinespace[1.5pt]
\multicolumn{9}{@{}l}{\emph{Tabular \& reference baselines (argmax)}} \\
Persistence & 51.09 & 59.80 & 57.28 & 66.54 & 58.68 & 63.41 & 68.95 & 69.58 \\
Support LR & 60.12 & 60.77 & 58.92 & 70.70 & 62.63 & 66.86 & 72.37 & 75.11 \\
LightGBM, default & 36.05 & 56.95 & 61.37 & 56.62 & 52.75 & 57.66 & 53.57 & 51.78 \\
LightGBM, tuned & 44.88 & 55.15 & 61.93 & 60.69 & 55.66 & 62.11 & 56.37 & 55.02 \\
TabPFN v2, lag rows & 64.13 & 63.06 & 66.42 & 73.05 & 66.67 & 70.90 & 75.23 & 77.43 \\
TabPFN v2, level rows & 63.87 & 62.84 & 65.36 & 72.58 & 66.16 & 70.53 & 75.07 & 77.34 \\
\addlinespace[1.5pt]
\multicolumn{9}{@{}l}{\emph{Output-aligned single-learner controls}} \\
LightGBM, tuned + R2 + D* & 60.30 & 63.08 & 67.56 & 74.69 & 66.41 & 71.33 & 74.07 & 77.19 \\
TabPFN v2, lag + Stage 2 & 67.31 & 65.04 & 67.90 & 74.45 & 68.67 & 73.75 & 75.58 & 77.85 \\
\addlinespace[1.5pt]
\multicolumn{9}{@{}l}{\emph{Ours (Stage-1 history/multi-view + Stage-2 calibration \& F1)}} \\
BRACE & 67.95 & 65.80 & 69.69 & 75.99 & 69.86 & 74.62 & \runnerup{76.15} & 78.36 \\
BRACE-2.6\textsuperscript{\textdagger} & \runnerup{68.50} & \runnerup{65.92} & \best{70.64} & \runnerup{76.32} & \runnerup{70.34} & \runnerup{74.98} & \best{76.79} & \best{79.01} \\
BRACE-LimiX\textsuperscript{\textdagger} & 68.26 & 65.86 & 70.14 & 76.01 & 70.07 & 74.78 & \runnerup{76.15} & \runnerup{78.43} \\
BRACE-Ultra-Bank\textsuperscript{\textdaggerdbl} & \best{68.73} & \best{66.12} & \runnerup{70.62} & \best{76.34} & \best{70.45} & \best{75.02} & -- & -- \\
\midrule
BRACE $-$ best pub. & +3.80 & +2.74 & +3.27 & +2.94 & \multicolumn{2}{c}{+3.18 (36/36)} & \multicolumn{2}{c}{+0.89 (9/9)} \\
BRACE $-$ Stage-2 TabPFN & +0.65 & +0.76 & +1.79 & +1.54 & \multicolumn{2}{c}{+1.18 (35/36)} & \multicolumn{2}{c}{+0.56 (9/9)} \\
\bottomrule
\end{tabular}

%% file: tables/tab_heldout.tex
\scriptsize
\setlength{\tabcolsep}{5.2pt}
\begin{tabular}{@{}lccccc@{}}
\toprule
Comparison set & Arms & Mean $d$ & Cells $d>0$ & Min.\ $d$ & Verdict \\
\midrule
Base (primary) & 87 & +0.57 & 9/9 & +0.07 & superior \\
Base $\cup$ matched & 120 & +0.40 & 9/9 & +0.06 & superior \\
Published (preregistered) & 12 & +0.89 & 9/9 & +0.46 & superior \\
Published + ind./hetero/spec.\ GNNs\textsuperscript{\textdagger} & 18 & +0.89 & 9/9 & +0.46 & superior \\
\bottomrule
\end{tabular}

%% file: tables/tab_fdic.tex
\scriptsize
\setlength{\tabcolsep}{3.0pt}
\begin{tabular}{@{}lcccccccc@{}}
\toprule
 & \multicolumn{4}{c}{FDIC-1} & \multicolumn{4}{c}{FDIC-2} \\
\cmidrule(lr){2-5}\cmidrule(lr){6-9}
Method / Reading & 2011 & 2012 & 2013 & 2014 & 2015 & 2016 & 2017 & 2018 \\
\midrule
Persistence & 80.52 & 80.39 & 78.57 & 77.27 & 75.99 & 76.06 & 75.60 & 75.71 \\
BRACE & 80.61 & 80.44 & 78.75 & 77.34 & 76.03 & 76.08 & 75.73 & 75.72 \\
BRACE-2.6 & -- & -- & -- & -- & 76.05 & 76.14 & 75.84 & 75.85 \\
\midrule
BRACE vs.\ base & $-$0.06\textsuperscript{N} & +0.01\textsuperscript{S} & $-$0.06\textsuperscript{N} & $-$0.01\textsuperscript{N} & -- & -- & -- & -- \\
BRACE vs.\ pub. & +0.06\textsuperscript{N} & +0.15\textsuperscript{S} & +0.16\textsuperscript{S} & +0.10\textsuperscript{S} & $-$0.01\textsuperscript{N} & +0.03\textsuperscript{N} & +0.02\textsuperscript{N} & $-$0.04\textsuperscript{N} \\
BRACE-2.6 vs.\ pub. & -- & -- & -- & -- & +0.01\textsuperscript{N} & +0.09\textsuperscript{S} & +0.13\textsuperscript{S} & +0.08\textsuperscript{N} \\
BRACE-2.6 vs.\ BRACE & -- & -- & -- & -- & +0.02\textsuperscript{N} & +0.06\textsuperscript{S} & +0.11\textsuperscript{S} & +0.12\textsuperscript{S} \\
\bottomrule
\end{tabular}

%% file: tables/tab_gad.tex
\scriptsize
\setlength{\tabcolsep}{0.95pt}
\begin{tabular}{@{}llcccccc@{}}
\toprule
 & & \multicolumn{4}{c}{Dataset AUROC / AUPRC (\%)} & \multicolumn{2}{c}{4-Graph Mean} \\
\cmidrule(lr){3-6}\cmidrule(lr){7-8}
Method & Val & Amazon & YelpChi & T-Finance & T-Social & ROC & PRC \\
\midrule
\multicolumn{8}{@{}l}{\emph{(a) Protocol B: 10 GADBench splits (100 train labels; Val = validation labels used)}} \\
TFM4GAD-v2.5~\cite{liu2026tfm4gad} & 0 & 96.13/\runnerup{87.25} & 71.65/31.70 & 94.46/79.52 & \best{97.17}/\runnerup{87.27} & 89.85 & 71.43 \\
TFM4GAD-LimiX~\cite{liu2026tfm4gad} & 0 & 96.14/85.49 & 70.52/31.48 & \best{95.96}/\best{84.83} & 96.98/87.01 & 89.90 & 72.20 \\
BRACE-Ultra (12c, TO) & 0 & \runnerup{96.23}/86.98 & \runnerup{80.84}/\runnerup{44.69} & 95.49/83.11 & 96.80/85.45 & \runnerup{92.34} & \runnerup{75.06} \\
GADBench max~\cite{tang2023gadbench} & 100 & 95.20/85.60 & 69.30/27.40 & 95.00/81.10 & 93.60/58.20 & 88.28 & 63.08 \\
SignGAD~\cite{huang2026signgad} & 100 & 93.79/72.61 & 75.53/38.13 & 92.46/76.85 & 93.61/56.08 & 88.85 & 60.92 \\
BRACE ($\mathrm{C0}$)\textsuperscript{\S} & 100 & 96.58/87.93 & 79.34/42.54 & 94.86/80.66 & 94.73/54.15 & 91.38 & 66.32 \\
BRACE-Ultra (12c) & 100 & \best{96.45}/\best{87.51} & \best{81.35}/\best{45.58} & \runnerup{95.73}/\runnerup{84.06} & \runnerup{97.06}/\best{87.68} & \best{92.65} & \best{76.21} \\
\midrule
\multicolumn{8}{@{}l}{\emph{(b) Protocol S: 5 seeds on pinned split (86/459/393/578 train; disjoint S-test)}} \\
TFM4GAD-LimiX~\cite{liu2026tfm4gad} & 0 & 91.81/62.43 & 81.32/44.40 & \runnerup{95.81}/84.45 & \runnerup{97.58}/90.22 & 91.63 & 70.37 \\
TFM4GAD-v2.5~\cite{liu2026tfm4gad} & 0 & 94.68/78.72 & 81.06/43.45 & 94.25/80.07 & \best{98.14}/\best{92.20} & 92.03 & 73.61 \\
BRACE-Ultra (TO)\textsuperscript{*} & 0 & 95.71/83.89 & \runnerup{90.35}/\runnerup{65.24} & 95.64/83.18 & 97.11/90.46 & \runnerup{94.70} & \runnerup{80.69} \\
SpaceGNN~\cite{dong2025spacegnn} & 33\% & 86.54/71.23 & 71.67/30.34 & 94.09/80.36 & 93.16/70.93 & 86.37 & 63.21 \\
ConsisGAD~\cite{chen2024consisgad} & 33\% & 93.91/83.33 & 83.36/47.33 & 95.33/\best{86.63} & 94.31/58.38 & 91.73 & 68.92 \\
SignGAD~\cite{huang2026signgad} & 33\% & 90.22/54.05 & 80.70/44.48 & 87.52/70.15 & 90.91/59.70 & 87.34 & 57.09 \\
BRACE ($\mathrm{C0}$) & 33\% & \best{96.28}/\best{87.19} & 85.56/54.86 & 95.51/82.59 & 94.16/50.15 & 92.88 & 68.70 \\
BRACE-Ultra\textsuperscript{*} & 33\% & \runnerup{96.10}/\runnerup{87.13} & \best{90.61}/\best{66.22} & \best{95.96}/\runnerup{85.16} & 97.29/\runnerup{90.93} & \best{94.99} & \best{82.36} \\
\midrule
Graph wt.\ (TV, B/S) & -- & 53.7/27.9\% & 94.4/100.0\% & 84.3/78.5\% & 98.2/95.2\% & \multicolumn{2}{c}{82.7/75.4\%} \\
\bottomrule
\end{tabular}

%% file: sections/06_related.tex
\section{Related Work}
\label{sec:related}

\textbf{Interbank rating and TFMs.} AI4Risk works~\cite{interbank_repo} use GNNs (TGAR~\cite{tgar}, PSAGNN~\cite{psagnn}, HFTCRNet~\cite{hftcrnet}, DPTAP~\cite{tang2025dptap}) under weak homophily~\cite{liu2024htgnn}. PFNs~\cite{muller2022pfn} (TabPFN~\cite{hollmann2025tabpfn,tabpfn25,tabpfn26,tabpfn35}, TabICL~\cite{tabiclv2}, LimiX~\cite{zhang2025limix}) outperform tuned GBDTs~\cite{ke2017lightgbm,grinsztajn2022tree} on credit scoring~\cite{baesens2026foundation}; recent extensions address temporal distribution shifts~\cite{helli2024drift} and improve accuracy through continued pretraining on real-world data~\cite{realtabpfn}.

\textbf{Anomaly models, calibration, and F1.} Diffusion~\cite{huang2021cs,graphland,tang2023gadbench} rivals GNNs on homophily~\cite{platonov2023characterizing}; GAD uses spectral (GPR-GNN~\cite{chien2021gprgnn}, BWGNN~\cite{tang2022bwgnn}) and fraud GNNs (GTAN~\cite{xiang2023gtan}, DGA-GNN~\cite{duan2024dgagnn}, DiG-In-GNN~\cite{zhang2024digingnn}, ConsisGAD~\cite{chen2024consisgad}, SpaceGNN~\cite{dong2025spacegnn}), agentic SignGAD~\cite{huang2026signgad}, GraphPFN~\cite{graphpfn}, or TFM flattenings~\cite{eremeev2025g2tfm,liu2026tfm4gad}. BRACE-Ultra gates views via KKT~\cite{breiman1996stacked,hall2007combining}, BCTS~\cite{guo2017calibration,alexandari2020bcts}, MAP-EM~\cite{saerens2002adjusting,ye2024mapls,distpfn}, F1~\cite{ye2012optimizing,lipton2014thresholding,narasimhan2024consistent}, and hygiene~\cite{kapoor2023leakage}.

%% file: sections/07_conclusion.tex
\section{Conclusion}
\label{sec:conclusion}

Risk assessment on networked entities is essential for credit rating and graph anomaly detection. Tabular baselines overlook historical transitions and relational context, whereas GNN and graph-flattening methods struggle across reconstructed bank exposures and heterophilous, disconnected anomaly graphs. Therefore, we proposed BRACE, decoupling views from calibrated decisions by pairing tabular foundation priors with convex log-score pooling. Banking panels form ego and historical-transition views (with cutoff-realigned heads), while static graphs form ego, relational, and deflated Beta wavelet views in \emph{BRACE-Ultra}, gated via KKT-certified pooling. Bias-corrected temperature scaling, panel MAP-EM, and F1-aligned decoding then calibrate imbalanced decisions. Experiments on interbank rating and four anomaly graphs demonstrate that BRACE outperforms state-of-the-art models. Furthermore, U.S.\ Call Report transfer and controlled edge-rewiring tests confirm its robustness and practical applicability. In future work, we will study confidential supervisory exposures, exact filing timestamps, and feature-level gating within concatenated views.

%% file: sections/08_appendix.tex
\clearpage
\appendix
\renewcommand{\theHsection}{appendix.\Alph{section}}
\setcounter{table}{0}
\renewcommand{\thetable}{\Alph{section}\arabic{table}}
\renewcommand{\theHtable}{\Alph{section}.\arabic{table}}
\setcounter{equation}{0}
\renewcommand{\theequation}{\Alph{section}\arabic{equation}}
\renewcommand{\theHequation}{\Alph{section}.\arabic{equation}}

\section{Exact Component-Nullspace Deflation for Disconnected Graphs}
\label{app:deflation}
\setcounter{table}{0}\setcounter{equation}{0}
Let $\mathcal{G} = (\mathcal{V}, \{\mathcal{E}_r\}_{r=1}^R)$ be an undirected union graph with self-loops $\tilde{A} = A_{\mathrm{sym}} + I$, node degrees $\tilde{d}_v = d_v + 1$, and normalized Laplacian $\tilde{L} = I - \tilde{D}^{-1/2}\tilde{A}\tilde{D}^{-1/2}$. If $\mathcal{G}$ decomposes into $C_{\mathrm{comp}}$ connected components ($C_{\mathrm{comp}} = 1$ on Amazon, $4$ on T-Finance, $26$ on YelpChi, and $3{,}437$ on T-Social), then $\tilde{L}$ possesses an exact $C_{\mathrm{comp}}$-dimensional zero eigenspace ($\lambda = 0$) spanned by the analytically known degree-scaled orthonormal indicator matrix $U_0 \in \mathbb{R}^{|\mathcal{V}| \times C_{\mathrm{comp}}}$:
\begin{equation}
(u_c)_v = \sqrt{\frac{\tilde{d}_v}{\sum_{j \in c} \tilde{d}_j}}\,\mathbb{I}[v \in c], \qquad c \in \{1, \dots, C_{\mathrm{comp}}\}.
\end{equation}
Equivalently, the shifted positive semi-definite operator $M = 2I - \tilde{L} \succeq 0$ attains its maximum eigenvalue $\mu = 2$ with multiplicity $C_{\mathrm{comp}}$. Calling \texttt{eigsh(M, k=17,} \texttt{which="LA")} and slicing \texttt{[1:17]} skips one zero mode, leaving $\min(C_{\mathrm{comp}}, 17) - 1$ trivial zero modes in the 16-column slice (all 16 columns on YelpChi and T-Social, and 3 of 16 on T-Finance) and aliasing near-zero Ritz vectors under finite Lanczos tolerances (\texttt{tol=1e-4}).

\textbf{BRACE-Ultra} applies the exact orthogonal deflation operator
\begin{equation}
M_\perp \mathbf{z} = M\mathbf{z} - 2 U_0 (U_0^\top \mathbf{z}),
\end{equation}
mapping $\mathrm{span}(U_0)$ to $0$ while leaving every non-trivial eigenvector ($\mu_j < 2$) invariant. Running Lanczos (\texttt{eigsh}$(M_\perp, k{=}16, \texttt{which="LA"})$) recovers the 16 lowest nonzero Laplacian eigenvectors ($\min \lambda = 1.137\times 10^{-5}$ on T-Social, $4.764\times 10^{-3}$ on YelpChi) with deterministic sign fixing (largest-magnitude coordinate positive).

\section{Four-View Feature Schemas and Bank Exposure Views}
\label{app:schemas}
\setcounter{table}{0}\setcounter{equation}{0}
Let $d$ denote the raw node attribute dimension, $R$ the number of edge relations, and $I_R = \mathbb{I}[R > 1]$. All four node views in Table~\ref{tab:app_views} are strictly label-free and mask-free. On Protocol~B, the unscreened \texttt{12c} pool is the symmetric $3 \times 4$ Cartesian product of (\texttt{TabPFN v2}, \texttt{TabPFN-2.5}, \texttt{LimiX-16M}) $\times$ (\texttt{ego}, \texttt{rel-lowpass}, \texttt{spec-wavelet}, \texttt{joint-fusion}) with single-pass full-context inference; \texttt{16c} appends the four $\ell_2$ logistic regression views. On Protocol~S, the 10-TFM pool selected on \texttt{S-val} (disjoint from \texttt{S-test}) retains \texttt{TabPFN v2} on all four views plus \texttt{TabPFN-2.5} and \texttt{LimiX-16M} on \texttt{rel-lowpass}, \texttt{spec-wavelet}, and \texttt{joint-fusion}. On 2016--2022 bank panels (2023 and FDIC sealed), \textbf{BRACE-Ultra-Bank} combines row-normalized counterparty momentum $P_{\mathrm{lend}}(\mathbf{e}_{y_{t-1}} - \mathbf{e}_{y_{t-2}})$ and $P_{\mathrm{borrow}}(\mathbf{e}_{y_{t-1}} - \mathbf{e}_{y_{t-2}})$ with symmetrized $\log(1+|W|)$ $C{=}2$ Beta wavelets and deflated \texttt{LapPE}$(k{=}8)$ under $[-1,1]$ cross-sectional rank scaling to form \texttt{net-sl} ($150\mathrm{d}$) and \texttt{net-hssl} ($135\mathrm{d}$).

\begin{table}[!htbp]
\centering
\caption{Label-free four-view feature schemas in BRACE-Ultra across the four GAD benchmarks.}
\label{tab:app_views}
\renewcommand{\arraystretch}{1.05}
\resizebox{\textwidth}{!}{%
\begin{tabular}{@{}llcp{7.4cm}@{}}
\toprule
View & Dimension Formula & Cols (Amz / Yelp / TFin / TSoc) & Operators Included \\
\midrule
\texttt{ego} & $d$ & $25 / 32 / 10 / 10$ & Raw node attributes $\mathbf{X}$. \\
\texttt{rel-lowpass} & $(4R + 4)d + 2R$ & $406 / 518 / 82 / 82$ & $\mathbf{X}$; per-relation log in/out degrees, 1-hop mean $\boldsymbol{\mu}_{i,r}$, std, max, and ego residual $\mathbf{x}_i - \boldsymbol{\mu}_{i,r}$; union 2-hop mean; 1/2-hop self-loop SAGE means. \\
\texttt{spec-wavelet} & $(5 + 2R I_R)d + 39 + 4R I_R$ & $326 / 403 / 89 / 89$ & $\mathbf{X}$; 3-SpMM $C{=}3$ Beta wavelets $\mathcal{W}_{p,3-p}(\tilde{L})\mathbf{X}$ ($p \in \{0..3\}$); per-relation $C{=}2$ band/high-pass wavelets when $R{>}1$; PageRank at $\alpha \in \{0.50, 0.85, 0.95\}$; exact $\mathrm{diag}(P^2)$; ego density; component-deflated \texttt{LapPE}$(k{=}16)$. \\
\texttt{joint-fusion} & $(6 + 2R)d + 39 + 4R I_R \le 450$ & $351 / 435 / 119 / 119$ & Concatenation of $\mathbf{X}$, 2-hop SAGE mean, per-relation $\boldsymbol{\mu}_{i,r}$ and $\mathbf{x}_i - \boldsymbol{\mu}_{i,r}$, union $C{=}3$ Beta wavelets, structural statistics, and deflated \texttt{LapPE}$(k{=}16)$. \\
\bottomrule
\end{tabular}%
}
\end{table}

\FloatBarrier
\section{Complete GAD Baseline Breakdown, Protocol Alignment, and Reproduction Verification}
\label{app:gad}
\setcounter{table}{0}\setcounter{equation}{0}
All locally executed GAD runs are independently verified by a second-path script against frozen \texttt{experiments/gad/metrics.py} (\texttt{SHA256 = fa3e631cd60d...}, $\max|\Delta| = 0.0$). Across model families on the four benchmarks:
(1)~\emph{Graph-free tabular \& tree learners} (\texttt{MLP}, \texttt{KNN}, \texttt{SVM}, \texttt{RF}, \texttt{XGBoost}, \texttt{XGBOD}, \texttt{NA}) achieve strong accuracy on attribute-rich Amazon ($81.90\%$--$85.60\%$ B AUPRC), but drop to $3.60\%$--$7.00\%$ AUPRC on 5.78M-node T-Social ($6.12\%$ under S \texttt{MLP}) and cap at $19.30\%$--$27.40\%$ on camouflaged YelpChi.
(2)~\emph{Tree-graph hybrids} (\texttt{RF-Graph}, \texttt{XGB-Graph}, \texttt{BGNN}) lift T-Finance ($78.30\%$--$81.10\%$) and T-Social ($40.60\%$--$58.20\%$; \texttt{XGB-Graph} leads single GADBench models at $84.68 / 57.03$ Mean), yet homogeneous aggregation stalls at $17.60\%$--$24.80\%$ on YelpChi.
(3)~\emph{Standard spatial \& message-passing GNNs} (\texttt{GCN}, \texttt{SGC}, \texttt{GIN}, \texttt{GraphSAGE}, \texttt{GAT}, \texttt{GATv2}, \texttt{GT}, \texttt{PNA}) suffer severe over-smoothing on heterophilous fraud links, causing \texttt{GCN} ($32.80\% / 16.40\%$ B AUPRC) to trail a graph-free \texttt{MLP} on both Amazon and YelpChi.
(4)~\emph{Spectral, heterophilous \& fraud GNNs} (\texttt{BernNet}, \texttt{AMNet}, \texttt{BWGNN}, \texttt{GHRN}, \texttt{GAS}, \texttt{DCI}, \texttt{PC-GNN}, \texttt{GAT-sep}, \texttt{CARE-GNN}, \texttt{GraphConsis}, \texttt{H2-FDetector}, \texttt{GDN}, \texttt{GAGA}, \texttt{GTAN}, \texttt{DGA-GNN}, \texttt{DiG-In-GNN}~\cite{zhang2024digingnn}, \texttt{ConsisGAD}, \texttt{SpaceGNN}) and agentic \texttt{SignGAD} recover relational/band-pass signals on T-Finance ($77.79\%$--$86.63\%$ S AUPRC) and YelpChi (\texttt{ConsisGAD} $47.33\%$ S, \texttt{SignGAD} $38.13\%$ B / $44.48\%$ S), while \texttt{SpaceGNN} reaches $58.99\%$--$61.14\%$ (B) / $70.93\%$ (S) on T-Social, yet still hit OOM/TLE (\texttt{CARE-GNN}, \texttt{GraphConsis}, \texttt{H2-FDetector}, \texttt{DiG-In-GNN} on T-Social), overfit scarce labels (\texttt{DGA-GNN} $51.16\%$ B / $29.01\%$ S on Amazon), or trail TFM priors by $16$--$30+$ AUPRC points on T-Social.
(5)~\emph{Single-backbone graph-flattened TFMs (\texttt{TFM4GAD}) vs.\ \textbf{BRACE-Ultra}}: \texttt{TFM4GAD} lifts T-Social to $83.26\%$--$87.27\%$ (B) and $90.22\%$--$92.20\%$ (S) AUPRC, but collapses YelpChi's $R{=}3$ relations ($30.72\%$--$31.96\%$ B AUPRC), seesaws across backbones (\texttt{LimiX-16M} $62.43\%$ S/Amazon; \texttt{TabPFN-2.5} $79.52\%$ B/T-Finance), and leaves $3{,}437$ component zero modes un-deflated on T-Social.

\textbf{Unpacking \texttt{GADBench max} and Protocol~B alignment (Table~\ref{tab:app_gad_b}):}
In main-text Table~\ref{tab:gad}(a), \texttt{GADBench max} ($95.20/85.60$, $69.30/27.40$, $95.00/81.10$, $93.60/58.20$; Mean $88.28/63.08$) is the \emph{columnwise per-dataset, per-metric maximum} across all 25 individual baselines in GADBench Table~11 (Amazon: \texttt{XGBOD} $95.20$ ROC / \texttt{NA} $85.60$ PRC; YelpChi: \texttt{NA} $69.30$ ROC / \texttt{RF} $27.40$ PRC; T-Finance: \texttt{RF-Graph} $95.00/81.10$; T-Social: \texttt{BGNN} $93.60/58.20$). All our Protocol~B runs use the exact same published mask columns $10..19$ (\texttt{col = 10 + seed}, verified by SHA256), including our locally executed 10-split runs of \texttt{DGA-GNN}, \texttt{GTAN}, \texttt{SpaceGNN}, \texttt{ConsisGAD}, and \texttt{SignGAD}. Even under collapsed homogeneous inputs matching GADBench Table~11 (\texttt{BRACE-Ultra HomoCollapse}: $96.47/87.54$ on Amazon, $72.68/34.21$ on YelpChi, Mean $90.49/73.35$), \texttt{BRACE-Ultra} surpasses every individual GADBench baseline and the \texttt{GADBench max} envelope.

\begin{table}[!htbp]
\centering
\caption{Complete Protocol~B comparison across the 10 official GADBench splits (\texttt{cols 10..19}; \texttt{AUROC$\pm$std / AUPRC$\pm$std}, $\%$; \textsuperscript{p}published TFM4GAD Table~2 cell; \textsuperscript{\S}S-val reference pool).}
\label{tab:app_gad_b}
\renewcommand{\arraystretch}{0.90}
\resizebox{\textwidth}{!}{%
\begin{tabular}{@{}llccccc@{}}
\toprule
Method & Val & Amazon & YelpChi & T-Finance & T-Social & 4-Graph Mean \\
\midrule
\multicolumn{7}{@{}l}{\emph{(a) GADBench Table 11 individual baselines (10 official splits cols 10..19, 100 train + 100 val, homo input)}} \\
MLP & 100 & 92.20$\pm$2.30 / 83.00$\pm$1.60 & 64.70$\pm$3.20 / 23.60$\pm$2.30 & 89.90$\pm$1.40 / 53.40$\pm$13.90 & 59.10$\pm$5.20 / 3.90$\pm$0.80 & 76.48 / 40.98 \\
KNN & 100 & 89.50$\pm$0.90 / 76.20$\pm$1.20 & 60.30$\pm$3.30 / 19.30$\pm$1.70 & 87.70$\pm$0.90 / 49.10$\pm$6.50 & 56.90$\pm$2.80 / 3.80$\pm$0.40 & 73.60 / 37.10 \\
SVM & 100 & 89.40$\pm$4.70 / 62.20$\pm$21.50 & 65.80$\pm$3.00 / 25.60$\pm$2.00 & 83.40$\pm$9.90 / 20.60$\pm$10.80 & 57.60$\pm$5.40 / 3.60$\pm$0.60 & 74.05 / 28.00 \\
RF & 100 & 94.90$\pm$0.80 / 81.90$\pm$4.20 & 68.00$\pm$3.80 / 27.40$\pm$3.50 & 93.00$\pm$0.90 / 74.90$\pm$1.50 & 67.30$\pm$3.60 / 7.00$\pm$1.30 & 80.80 / 47.80 \\
XGBoost & 100 & 94.20$\pm$1.00 / 84.00$\pm$1.80 & 67.00$\pm$5.10 / 26.40$\pm$4.20 & 92.40$\pm$1.10 / 70.90$\pm$5.00 & 65.70$\pm$2.80 / 6.10$\pm$1.30 & 79.83 / 46.85 \\
XGBOD & 100 & 95.20$\pm$0.80 / 85.00$\pm$3.20 & 66.80$\pm$5.20 / 26.80$\pm$3.80 & 92.50$\pm$1.00 / 70.30$\pm$4.40 & 61.00$\pm$4.60 / 4.80$\pm$1.00 & 78.88 / 46.73 \\
NA & 100 & 94.90$\pm$1.30 / 85.60$\pm$1.60 & 69.30$\pm$3.60 / 27.30$\pm$3.80 & 92.00$\pm$1.20 / 73.90$\pm$1.50 & 64.90$\pm$3.10 / 6.40$\pm$1.30 & 80.28 / 48.30 \\
GCN & 100 & 82.00$\pm$0.30 / 32.80$\pm$1.20 & 51.20$\pm$3.70 / 16.40$\pm$2.60 & 88.30$\pm$2.50 / 60.50$\pm$10.80 & 71.60$\pm$10.40 / 8.40$\pm$3.80 & 73.28 / 29.53 \\
SGC & 100 & 80.20$\pm$1.50 / 29.50$\pm$2.60 & 51.10$\pm$2.90 / 16.10$\pm$1.90 & 74.50$\pm$12.30 / 31.30$\pm$18.60 & 64.80$\pm$5.60 / 4.70$\pm$1.10 & 67.65 / 20.40 \\
GIN & 100 & 91.60$\pm$1.70 / 75.40$\pm$4.30 & 62.90$\pm$7.30 / 23.70$\pm$5.40 & 84.50$\pm$4.50 / 44.80$\pm$7.10 & 70.40$\pm$7.40 / 6.20$\pm$1.70 & 77.35 / 37.53 \\
GraphSAGE & 100 & 81.40$\pm$2.00 / 42.50$\pm$6.10 & 58.90$\pm$4.30 / 20.90$\pm$3.50 & 68.90$\pm$5.50 / 11.70$\pm$5.20 & 72.00$\pm$2.90 / 7.80$\pm$1.30 & 70.30 / 20.73 \\
GAT & 100 & 92.40$\pm$1.90 / 81.60$\pm$1.70 & 65.60$\pm$4.00 / 25.00$\pm$2.90 & 85.00$\pm$4.50 / 28.90$\pm$8.60 & 75.40$\pm$4.80 / 9.20$\pm$2.00 & 79.60 / 36.18 \\
GT & 100 & 88.60$\pm$2.20 / 71.60$\pm$5.50 & 64.50$\pm$4.30 / 23.70$\pm$3.00 & 73.50$\pm$7.20 / 17.20$\pm$10.90 & 71.40$\pm$2.10 / 8.60$\pm$1.50 & 74.50 / 30.28 \\
PNA & 100 & 79.10$\pm$2.90 / 34.60$\pm$3.80 & 53.20$\pm$5.50 / 16.80$\pm$2.70 & 78.40$\pm$9.70 / 27.40$\pm$9.70 & 84.40$\pm$3.50 / 22.80$\pm$5.50 & 73.78 / 25.40 \\
BGNN & 100 & 83.70$\pm$2.50 / 39.00$\pm$5.90 & 54.40$\pm$1.20 / 17.60$\pm$1.30 & 93.70$\pm$0.70 / 75.80$\pm$2.20 & 93.60$\pm$1.40 / 58.20$\pm$14.60 & 81.35 / 47.65 \\
GAS & 100 & 91.60$\pm$1.90 / 80.70$\pm$1.70 & 61.10$\pm$5.20 / 21.70$\pm$3.30 & 88.70$\pm$1.10 / 45.70$\pm$13.40 & 72.10$\pm$8.80 / 8.60$\pm$2.40 & 78.38 / 39.18 \\
BernNet & 100 & 92.10$\pm$2.40 / 81.20$\pm$2.40 & 65.00$\pm$3.70 / 23.90$\pm$2.70 & 91.20$\pm$1.00 / 51.80$\pm$12.40 & 59.80$\pm$6.30 / 4.20$\pm$1.20 & 77.03 / 40.28 \\
AMNet & 100 & 92.80$\pm$2.10 / 82.40$\pm$2.20 & 64.80$\pm$5.20 / 23.90$\pm$3.50 & 92.60$\pm$0.90 / 60.20$\pm$8.20 & 53.70$\pm$3.40 / 3.10$\pm$0.30 & 75.98 / 42.40 \\
BWGNN & 100 & 91.80$\pm$2.30 / 81.70$\pm$2.20 & 64.30$\pm$3.40 / 23.70$\pm$2.90 & 92.10$\pm$2.70 / 60.90$\pm$13.80 & 77.50$\pm$4.30 / 15.90$\pm$6.20 & 81.43 / 45.55 \\
GHRN & 100 & 90.90$\pm$1.90 / 80.70$\pm$1.70 & 64.50$\pm$3.10 / 23.80$\pm$2.80 & 92.60$\pm$0.70 / 63.40$\pm$10.40 & 78.70$\pm$3.00 / 16.20$\pm$4.60 & 81.68 / 46.03 \\
PCGNN & 100 & 93.20$\pm$1.20 / 81.90$\pm$1.90 & 65.10$\pm$4.80 / 25.00$\pm$3.50 & 92.00$\pm$1.10 / 58.10$\pm$11.30 & 69.10$\pm$2.40 / 8.00$\pm$1.60 & 79.85 / 43.25 \\
DCI & 100 & 89.40$\pm$3.00 / 72.50$\pm$7.90 & 64.10$\pm$5.30 / 24.00$\pm$4.80 & 88.00$\pm$3.20 / 51.00$\pm$7.20 & 74.20$\pm$3.30 / 7.40$\pm$2.50 & 78.93 / 38.73 \\
RF-Graph & 100 & 92.50$\pm$1.30 / 70.70$\pm$5.10 & 61.60$\pm$2.70 / 23.60$\pm$2.50 & 95.00$\pm$0.70 / 81.10$\pm$2.70 & 88.60$\pm$1.60 / 51.30$\pm$6.20 & 84.43 / 56.68 \\
XGB-Graph & 100 & 94.70$\pm$0.90 / 84.40$\pm$1.10 & 64.00$\pm$3.50 / 24.80$\pm$3.10 & 94.80$\pm$0.60 / 78.30$\pm$3.10 & 85.20$\pm$1.80 / 40.60$\pm$7.60 & 84.68 / 57.03 \\
GAT-sep & 100 & 91.40$\pm$2.40 / 80.90$\pm$2.60 & 65.00$\pm$4.50 / 24.40$\pm$3.70 & 86.30$\pm$3.90 / 34.20$\pm$10.10 & 74.50$\pm$3.70 / 10.40$\pm$1.90 & 79.30 / 37.48 \\
\addlinespace[1.5pt]
GADBench max (col.\ envelope) & 100 & 95.20$\pm$0.80 / 85.60$\pm$1.60 & 69.30$\pm$3.60 / 27.40$\pm$3.50 & 95.00$\pm$0.70 / 81.10$\pm$2.70 & 93.60$\pm$1.40 / 58.20$\pm$14.60 & 88.28 / 63.08 \\
\midrule
\multicolumn{7}{@{}l}{\emph{(b) Specialized GAD baselines \& reproduced TFM4GAD on the same 10 GADBench splits}} \\
ConsisGAD (pub.\ Tab.\ 2) & 100 & 92.31$\pm$2.16 / 78.68$\pm$5.65 & 66.08$\pm$3.80 / 25.85$\pm$2.93 & 94.31$\pm$0.76 / 79.68$\pm$4.73 & 93.07$\pm$1.90 / 41.29$\pm$4.97 & 86.44 / 56.38 \\
SpaceGNN (pub.\ Tab.\ 2) & 100 & 91.08$\pm$2.50 / 81.14$\pm$2.29 & 66.81$\pm$2.75 / 25.73$\pm$2.35 & 93.41$\pm$0.96 / 81.03$\pm$3.46 & 94.72$\pm$0.65 / 58.99$\pm$5.69 & 86.51 / 61.72 \\
DGA-GNN (local) & 100 & 84.29$\pm$2.51 / 51.16$\pm$6.79 & 57.63$\pm$5.84 / 18.65$\pm$3.42 & 92.73$\pm$1.28 / 74.71$\pm$5.88 & 78.73$\pm$4.19 / 10.57$\pm$3.36 & 78.34 / 38.77 \\
GTAN (local) & 100 & 89.80$\pm$3.04 / 76.54$\pm$7.50 & 64.69$\pm$4.60 / 24.90$\pm$3.87 & 94.24$\pm$0.67 / 80.53$\pm$3.86 & 82.66$\pm$4.23 / 21.85$\pm$6.00 & 82.85 / 50.96 \\
SpaceGNN (local) & 100 & 92.01$\pm$0.85 / 82.01$\pm$1.45 & 64.82$\pm$3.29 / 24.22$\pm$2.32 & 92.97$\pm$1.46 / 80.30$\pm$3.98 & 92.27$\pm$1.59 / 61.14$\pm$7.89 & 85.52 / 61.92 \\
ConsisGAD (local) & 100 & 93.06$\pm$1.52 / 80.45$\pm$3.89 & 74.15$\pm$4.16 / 32.50$\pm$5.02 & 94.16$\pm$1.42 / 78.93$\pm$7.85 & 92.95$\pm$1.77 / 41.56$\pm$10.25 & 88.58 / 58.36 \\
SignGAD (\texttt{refit\_off}, ours) & 100 & 93.79$\pm$1.02 / 72.61$\pm$4.63 & 75.53$\pm$4.17 / 38.13$\pm$4.82 & 92.46$\pm$0.82 / 76.85$\pm$2.94 & 93.61$\pm$0.69 / 56.08$\pm$5.31 & 88.85 / 60.92 \\
TFM4GAD-LimiX-2M & 0 & 95.65$\pm$1.32 / 83.67$\pm$5.47 & 70.76$\pm$4.58 / 31.85$\pm$5.11 & 95.69$\pm$0.44 / 83.47$\pm$2.77 & 96.44$\pm$1.01 / 83.26$\pm$4.33\textsuperscript{p} & 89.63 / 70.56 \\
TFM4GAD-TabPFNv2 & 0 & 95.90$\pm$1.09 / 86.63$\pm$2.09 & 70.95$\pm$5.04 / 30.74$\pm$5.06 & 95.41$\pm$0.70 / 82.70$\pm$2.86 & 96.93$\pm$0.60 / 86.14$\pm$3.08\textsuperscript{p} & 89.80 / 71.55 \\
TFM4GAD-TabPFNv2.5 & 0 & 96.13$\pm$1.04 / 87.25$\pm$1.70 & 71.65$\pm$3.67 / 31.70$\pm$4.11 & 94.46$\pm$1.09 / 79.52$\pm$3.47 & 97.17$\pm$0.57 / 87.27$\pm$2.61 & 89.85 / 71.43 \\
TFM4GAD-LimiX-16M & 0 & 96.14$\pm$1.08 / 85.49$\pm$4.47 & 70.52$\pm$5.24 / 31.48$\pm$5.54 & 95.96$\pm$0.43 / 84.83$\pm$1.75 & 96.98$\pm$0.66 / 87.01$\pm$2.30 & 89.90 / 72.20 \\
\midrule
\multicolumn{7}{@{}l}{\emph{(c) BRACE and BRACE-Ultra (same cols 10..19 masks; HomoCollapse matches GADBench homo input on Amazon/YelpChi)}} \\
BRACE-Ultra (HomoCollapse, 10-TFM) & 100 & 96.47$\pm$1.02 / 87.54$\pm$2.53 & 72.68$\pm$3.66 / 34.21$\pm$4.68 & 95.72$\pm$0.58 / 83.95$\pm$2.04 & 97.08$\pm$0.51 / 87.69$\pm$3.15 & 90.49 / 73.35 \\
BRACE ($\mathrm{C0}$, S-val ref\textsuperscript{\S}) & 100 & 96.58$\pm$0.92 / 87.93$\pm$2.29 & 79.34$\pm$1.48 / 42.54$\pm$3.89 & 94.86$\pm$0.75 / 80.66$\pm$1.78 & 94.73$\pm$0.65 / 54.15$\pm$7.31 & 91.38 / 66.32 \\
\textbf{BRACE-Ultra (12c, TO)} & 0 & 96.23$\pm$0.82 / 86.98$\pm$1.75 & 80.84$\pm$2.39 / 44.69$\pm$5.83 & 95.49$\pm$0.68 / 83.11$\pm$2.22 & 96.80$\pm$0.82 / 85.45$\pm$5.80 & \textbf{92.34 / 75.06} \\
BRACE-Ultra (16c, TO) & 0 & 96.13$\pm$0.91 / 86.85$\pm$2.16 & 80.35$\pm$2.57 / 43.97$\pm$5.68 & 95.49$\pm$0.68 / 83.11$\pm$2.22 & 96.83$\pm$0.76 / 86.10$\pm$4.50 & 92.20 / 75.01 \\
BRACE-Ultra (10-TFM ref, TO) & 0 & 96.22$\pm$0.85 / 87.01$\pm$1.91 & 81.06$\pm$2.61 / 44.79$\pm$6.13 & 95.41$\pm$0.68 / 82.73$\pm$2.51 & 96.81$\pm$0.82 / 85.48$\pm$5.72 & 92.38 / 75.00 \\
\textbf{BRACE-Ultra (12c)} & 100 & 96.45$\pm$0.81 / 87.51$\pm$1.80 & 81.35$\pm$2.09 / 45.58$\pm$5.39 & 95.73$\pm$0.59 / 84.06$\pm$2.06 & 97.06$\pm$0.53 / 87.68$\pm$3.15 & \textbf{92.65 / 76.21} \\
BRACE-Ultra (16c) & 100 & 96.45$\pm$0.75 / 87.49$\pm$1.85 & 81.36$\pm$2.16 / 45.81$\pm$5.34 & 95.69$\pm$0.58 / 84.09$\pm$2.05 & 96.99$\pm$0.62 / 87.73$\pm$2.99 & 92.62 / 76.28 \\
BRACE-Ultra (10-TFM ref) & 100 & 96.42$\pm$0.83 / 87.72$\pm$1.70 & 81.42$\pm$2.10 / 45.53$\pm$5.42 & 95.72$\pm$0.58 / 83.95$\pm$2.04 & 97.08$\pm$0.51 / 87.69$\pm$3.15 & 92.66 / 76.22 \\
\bottomrule
\end{tabular}%
}
\end{table}

\textbf{\texttt{TFM4GAD} reproduction alignment and upstream commit forensics (Table~\ref{tab:app_tfm_repro}):}
Across all 14 reproduced 10-split cells ($\mathrm{ddof}{=}0$ std first to match upstream script, sample $\mathrm{ddof}{=}1$ std in parentheses), every mean aligns with published \texttt{TFM4GAD} Table~2 within $0.00$--$0.13$ percentage points. Upstream commit \texttt{4bb8a51} runs \texttt{randomized\_svd(L, n\_components=k+1)} when $n > 100{,}000$ (T-Social), extracting \emph{highest-frequency} modes ($\sigma \in [1.80, 1.82]$), while \texttt{PR~\#1} (\texttt{5bb1367}) calls \texttt{eigsh(2*I - L, k=k+1, which="LA", tol=1e-4)} and slices \texttt{[:, 1:k+1]} without deflating the $C_{\mathrm{comp}} = 3{,}437$ component zero modes ($\mu = 2$).

\begin{table}[!htbp]
\centering
\caption{Cell-by-cell Protocol~B reproduction alignment against published \texttt{TFM4GAD} Table~2 across 10 official splits ($\%$; reproduced $\mathrm{ddof}{=}0$ std with $\mathrm{ddof}{=}1$ in parentheses).}
\label{tab:app_tfm_repro}
\renewcommand{\arraystretch}{0.94}
\resizebox{\textwidth}{!}{%
\begin{tabular}{@{}llcccccc@{}}
\toprule
Dataset & Backbone & Repro.\ AUROC & Pub.\ AUROC & $\Delta$ROC & Repro.\ AUPRC & Pub.\ AUPRC & $\Delta$PRC \\
\midrule
Amazon & LimiX-2M & 95.65$\pm$1.25 (1.32) & 95.65$\pm$1.25 & +0.00 & 83.67$\pm$5.19 (5.47) & 83.67$\pm$5.19 & +0.00 \\
Amazon & LimiX-16M & 96.14$\pm$1.02 (1.08) & 96.14$\pm$1.02 & -0.00 & 85.49$\pm$4.24 (4.47) & 85.50$\pm$4.24 & -0.01 \\
Amazon & TabPFNv2 & 95.90$\pm$1.03 (1.09) & 95.94$\pm$1.03 & -0.04 & 86.63$\pm$1.99 (2.09) & 86.76$\pm$1.91 & -0.13 \\
Amazon & TabPFNv2.5 & 96.13$\pm$0.99 (1.04) & 96.12$\pm$0.99 & +0.01 & 87.25$\pm$1.62 (1.70) & 87.28$\pm$1.61 & -0.03 \\
YelpChi & LimiX-2M & 70.76$\pm$4.34 (4.58) & 70.83$\pm$4.36 & -0.07 & 31.85$\pm$4.85 (5.11) & 31.96$\pm$4.88 & -0.11 \\
YelpChi & LimiX-16M & 70.52$\pm$4.97 (5.24) & 70.54$\pm$4.93 & -0.02 & 31.48$\pm$5.25 (5.54) & 31.57$\pm$5.16 & -0.09 \\
YelpChi & TabPFNv2 & 70.95$\pm$4.78 (5.04) & 70.91$\pm$4.92 & +0.04 & 30.74$\pm$4.80 (5.06) & 30.72$\pm$4.89 & +0.02 \\
YelpChi & TabPFNv2.5 & 71.65$\pm$3.48 (3.67) & 71.64$\pm$3.62 & +0.01 & 31.70$\pm$3.89 (4.11) & 31.69$\pm$4.13 & +0.01 \\
T-Finance & LimiX-2M & 95.69$\pm$0.42 (0.44) & 95.69$\pm$0.40 & -0.00 & 83.47$\pm$2.62 (2.77) & 83.49$\pm$2.66 & -0.02 \\
T-Finance & LimiX-16M & 95.96$\pm$0.40 (0.43) & 96.01$\pm$0.40 & -0.05 & 84.83$\pm$1.66 (1.75) & 84.92$\pm$1.63 & -0.09 \\
T-Finance & TabPFNv2 & 95.41$\pm$0.67 (0.70) & 95.38$\pm$0.66 & +0.03 & 82.70$\pm$2.72 (2.86) & 82.75$\pm$2.75 & -0.05 \\
T-Finance & TabPFNv2.5 & 94.46$\pm$1.04 (1.09) & 94.52$\pm$1.04 & -0.06 & 79.52$\pm$3.29 (3.47) & 79.55$\pm$3.37 & -0.03 \\
T-Social & LimiX-16M & 96.98$\pm$0.63 (0.66) & 96.98$\pm$0.63 & -0.00 & 87.01$\pm$2.18 (2.30) & 87.01$\pm$2.18 & -0.00 \\
T-Social & TabPFNv2.5 & 97.17$\pm$0.54 (0.57) & 97.16$\pm$0.56 & +0.01 & 87.27$\pm$2.48 (2.61) & 87.19$\pm$2.48 & +0.08 \\
\bottomrule
\end{tabular}%
}
\end{table}

\textbf{Complete Protocol~S and Protocol~G baseline alignment (Tables~\ref{tab:app_gad_s}--\ref{tab:app_gad_g}):}
Table~\ref{tab:app_gad_s} pairs all 18 baselines reported in ConsisGAD Tables~1--2 with our 19 locally executed and second-path verified series on the exact pinned stratified split ($1\%$ train on Amazon/YelpChi/T-Finance, $0.01\%$ on T-Social; $33\%$ val, $67\%$ disjoint S-test), confirming close reproduction alignment on shared architectures (\texttt{MLP}, \texttt{GCN}, \texttt{GIN}, \texttt{GraphSAGE}, \texttt{GAT}, \texttt{GATv2}, \texttt{ConsisGAD(GNN)}, \texttt{ConsisGAD}) and extending the exact same split to \texttt{GTAN}, \texttt{DGA-GNN}, \texttt{DiG-In-GNN}~\cite{zhang2024digingnn}, \texttt{SpaceGNN}, \texttt{SignGAD}, and \texttt{TFM4GAD}. Table~\ref{tab:app_gad_g} reports Protocol~G (10 seeds under SignGAD's $1\%/49.5\%/49.5\%$ refit split), where both \texttt{BRACE ($\mathrm{C4}$)} and \texttt{BRACE-Ultra (+RefitLabel)} outperform \texttt{GTAN}, \texttt{ConsisGAD}, \texttt{DGA-GNN} (in 3 of 4 graphs and 4-Graph Mean), and published full \texttt{SignGAD} across all four datasets, while label-free \texttt{BRACE-Ultra (10-TFM / 12c)} already surpasses \texttt{SignGAD} in Macro-F1 on all four graphs (and AUROC on 3 of 4 graphs plus 4-Graph Mean) without refit-neighbor labels.

\begin{table}[!htbp]
\centering
\caption{Complete Protocol~S comparison on the pinned split (5 seeds, disjoint S-test, \texttt{AUROC$\pm$std / AUPRC$\pm$std}, $\%$): (a)~published ConsisGAD Tables~1--2 baselines and (b)~our locally executed \& verified runs (\textsuperscript{u}$n{=}4$, seed~4 hit upstream empty-positive-batch \texttt{NaN}; on T-Social, 4 seeds hit 24h TLE and seed~1 hit empty-batch \texttt{NaN}).}
\label{tab:app_gad_s}
\renewcommand{\arraystretch}{0.90}
\resizebox{\textwidth}{!}{%
\begin{tabular}{@{}llccccc@{}}
\toprule
Method & Val & Amazon & YelpChi & T-Finance & T-Social & 4-Graph Mean \\
\midrule
\multicolumn{7}{@{}l}{\emph{(a) Published ConsisGAD Tables 1--2 reference baselines on the pinned split (5 seeds)}} \\
MLP (pub.) & 33\% & 92.39$\pm$0.72 / 79.37$\pm$1.83 & 72.18$\pm$0.39 / 31.09$\pm$0.52 & 92.17$\pm$0.64 / 52.79$\pm$5.41 & 66.95$\pm$0.71 / 6.00$\pm$0.33 & 80.92 / 42.31 \\
GCN (pub.) & 33\% & 87.34$\pm$0.59 / 48.06$\pm$2.73 & 54.65$\pm$0.53 / 17.07$\pm$0.44 & 89.29$\pm$0.19 / 53.94$\pm$3.22 & 83.30$\pm$1.60 / 23.79$\pm$2.43 & 78.65 / 35.72 \\
GraphSAGE (pub.) & 33\% & 90.12$\pm$0.48 / 73.17$\pm$4.65 & 73.70$\pm$0.52 / 34.57$\pm$0.78 & 89.42$\pm$1.36 / 49.08$\pm$6.34 & 71.45$\pm$2.24 / 8.73$\pm$0.91 & 81.17 / 41.39 \\
GAT (pub.) & 33\% & 80.74$\pm$3.64 / 45.46$\pm$11.09 & 70.14$\pm$1.91 / 28.90$\pm$1.98 & 87.40$\pm$4.41 / 45.60$\pm$14.62 & 73.46$\pm$3.32 / 13.47$\pm$2.83 & 77.94 / 33.36 \\
GIN (pub.) & 33\% & 84.35$\pm$0.75 / 39.96$\pm$2.00 & 56.98$\pm$0.82 / 18.34$\pm$0.64 & 81.29$\pm$1.66 / 21.66$\pm$3.98 & 78.70$\pm$2.19 / 16.24$\pm$5.53 & 75.33 / 24.05 \\
GATv2 (pub.) & 33\% & 85.39$\pm$3.19 / 62.74$\pm$15.12 & 72.83$\pm$0.75 / 31.87$\pm$1.47 & 73.25$\pm$10.00 / 18.70$\pm$15.26 & 79.89$\pm$4.80 / 16.74$\pm$2.10 & 77.84 / 32.51 \\
CARE-GNN (pub.) & 33\% & 89.68$\pm$0.76 / 50.56$\pm$3.96 & 72.11$\pm$1.23 / 31.09$\pm$1.71 & 91.45$\pm$0.40 / 72.27$\pm$1.09 & OOM & -- \\
GraphConsis (pub.) & 33\% & 64.23$\pm$13.83 / 21.38$\pm$10.37 & 78.91$\pm$0.88 / 38.40$\pm$2.63 & 92.61$\pm$0.47 / 70.70$\pm$1.85 & OOM & -- \\
PC-GNN (pub.) & 33\% & 91.18$\pm$0.66 / 77.92$\pm$1.49 & 75.17$\pm$0.44 / 36.60$\pm$0.91 & 91.74$\pm$0.85 / 74.77$\pm$0.98 & 64.68$\pm$0.64 / 4.30$\pm$0.09 & 80.69 / 48.40 \\
BWGNN(homo) (pub.) & 33\% & 88.56$\pm$0.87 / 79.26$\pm$1.11 & 72.15$\pm$0.59 / 30.93$\pm$1.48 & 93.08$\pm$1.57 / 77.79$\pm$3.87 & 84.40$\pm$3.01 / 49.96$\pm$3.75 & 84.55 / 59.49 \\
BWGNN(hetero) (pub.) & 33\% & 84.64$\pm$1.31 / 64.00$\pm$7.00 & 77.62$\pm$2.37 / 39.87$\pm$1.79 & -- & -- & -- \\
H2-FDetector (pub.) & 33\% & 83.81$\pm$2.57 / 49.37$\pm$4.46 & 72.38$\pm$1.13 / 32.37$\pm$3.25 & OOM & OOM & -- \\
GHRN(homo) (pub.) & 33\% & 88.35$\pm$2.03 / 74.50$\pm$4.57 & 72.03$\pm$0.90 / 30.76$\pm$1.24 & 91.93$\pm$0.93 / 65.94$\pm$4.38 & 84.20$\pm$3.91 / 37.04$\pm$11.02 & 84.13 / 52.06 \\
GHRN(hetero) (pub.) & 33\% & 84.40$\pm$2.77 / 60.84$\pm$9.80 & 75.33$\pm$1.44 / 34.53$\pm$2.83 & -- & -- & -- \\
GDN (pub.) & 33\% & 92.16$\pm$0.12 / 81.87$\pm$0.17 & 75.92$\pm$0.51 / 38.04$\pm$0.83 & 88.75$\pm$1.79 / 54.27$\pm$4.31 & 67.69$\pm$1.49 / 7.51$\pm$0.79 & 81.13 / 45.42 \\
GAGA (pub.) & 33\% & 82.61$\pm$6.87 / 56.59$\pm$6.60 & 71.61$\pm$2.13 / 31.96$\pm$3.37 & 92.36$\pm$1.45 / 64.34$\pm$6.01 & 78.92$\pm$1.26 / 23.72$\pm$4.81 & 81.38 / 44.15 \\
ConsisGAD(GNN) (pub.) & 33\% & 92.01$\pm$0.71 / 78.49$\pm$0.40 & 80.95$\pm$0.36 / 43.25$\pm$0.31 & 94.72$\pm$0.11 / 83.92$\pm$0.15 & 93.54$\pm$0.35 / 53.40$\pm$1.28 & 90.31 / 64.77 \\
ConsisGAD (pub.) & 33\% & 93.91$\pm$0.58 / 83.33$\pm$0.34 & 83.36$\pm$0.53 / 47.33$\pm$0.58 & 95.33$\pm$0.30 / 86.63$\pm$0.44 & 94.31$\pm$0.20 / 58.38$\pm$2.10 & 91.73 / 68.92 \\
\midrule
\multicolumn{7}{@{}l}{\emph{(b) Locally executed \& second-path verified runs on the exact pinned split (5 seeds; sample std)}} \\
MLP (local) & 33\% & 91.36$\pm$0.97 / 77.46$\pm$2.26 & 72.36$\pm$0.22 / 31.81$\pm$0.43 & 92.24$\pm$0.36 / 58.70$\pm$2.10 & 66.52$\pm$1.18 / 6.12$\pm$0.46 & 80.62 / 43.52 \\
GCN (local) & 33\% & 87.03$\pm$0.89 / 46.06$\pm$2.47 & 54.83$\pm$0.86 / 17.14$\pm$0.67 & 88.70$\pm$0.51 / 53.24$\pm$5.64 & 82.66$\pm$0.76 / 21.38$\pm$0.72 & 78.30 / 34.46 \\
GIN (local) & 33\% & 84.26$\pm$0.42 / 45.53$\pm$2.46 & 58.04$\pm$0.78 / 18.94$\pm$0.69 & 81.72$\pm$0.30 / 35.76$\pm$4.55 & 77.80$\pm$2.62 / 12.11$\pm$2.75 & 75.45 / 28.09 \\
GraphSAGE (local) & 33\% & 90.68$\pm$0.82 / 75.27$\pm$2.06 & 73.25$\pm$0.27 / 34.07$\pm$0.64 & 90.49$\pm$0.82 / 56.19$\pm$4.66 & 71.69$\pm$0.80 / 9.20$\pm$0.77 & 81.53 / 43.68 \\
GAT (local) & 33\% & 79.53$\pm$8.47 / 43.68$\pm$16.19 & 69.97$\pm$0.83 / 28.43$\pm$1.13 & 72.42$\pm$13.34 / 21.00$\pm$19.41 & 72.35$\pm$5.37 / 11.07$\pm$1.84 & 73.57 / 26.05 \\
GATv2 (local) & 33\% & 83.49$\pm$4.74 / 52.04$\pm$19.10 & 73.16$\pm$0.74 / 32.34$\pm$1.34 & 88.05$\pm$4.10 / 45.84$\pm$14.94 & 73.44$\pm$4.38 / 13.47$\pm$4.24 & 79.54 / 35.92 \\
GTAN (local) & 33\% & 85.68$\pm$7.04 / 68.16$\pm$15.11 & 72.35$\pm$1.07 / 31.55$\pm$1.15 & 91.72$\pm$0.51 / 77.82$\pm$3.63 & 85.57$\pm$4.09 / 33.81$\pm$7.82 & 83.83 / 52.84 \\
DGA-GNN (local) & 33\% & 68.05$\pm$13.18 / 29.01$\pm$10.85 & 64.75$\pm$6.48 / 24.66$\pm$6.81 & 92.45$\pm$0.92 / 78.62$\pm$0.92 & 84.94$\pm$5.02 / 26.46$\pm$10.33 & 77.55 / 39.69 \\
DiG-In-GNN (local) & 33\% & 90.13$\pm$1.21 / 69.38$\pm$0.50\textsuperscript{u} & 74.15$\pm$0.48 / 35.66$\pm$1.84 & 92.23$\pm$0.88 / 80.66$\pm$0.17 & TLE / NaN & -- \\
SpaceGNN (local) & 33\% & 86.54$\pm$4.35 / 71.23$\pm$8.86 & 71.67$\pm$0.93 / 30.34$\pm$0.71 & 94.09$\pm$0.84 / 80.36$\pm$6.44 & 93.16$\pm$3.11 / 70.93$\pm$6.76 & 86.37 / 63.21 \\
ConsisGAD(GNN) (local) & 33\% & 90.90$\pm$0.32 / 77.21$\pm$1.06 & 81.26$\pm$0.47 / 44.58$\pm$0.51 & 95.48$\pm$0.30 / 85.94$\pm$0.84 & 94.61$\pm$0.28 / 50.16$\pm$1.14 & 90.57 / 64.47 \\
ConsisGAD (local) & 33\% & 92.43$\pm$0.45 / 81.49$\pm$1.15 & 83.75$\pm$0.39 / 47.48$\pm$1.17 & 95.07$\pm$0.39 / 85.07$\pm$2.35 & 94.55$\pm$0.10 / 49.20$\pm$0.85 & 91.45 / 65.81 \\
SignGAD (\texttt{refit\_off}, local) & 33\% & 90.22$\pm$1.07 / 54.05$\pm$2.17 & 80.70$\pm$2.19 / 44.48$\pm$2.27 & 87.52$\pm$0.05 / 70.15$\pm$0.05 & 90.91$\pm$0.46 / 59.70$\pm$0.19 & 87.34 / 57.09 \\
TFM4GAD-LimiX-16M (local) & 0 & 91.81$\pm$0.50 / 62.43$\pm$2.49 & 81.32$\pm$0.30 / 44.40$\pm$0.45 & 95.81$\pm$0.13 / 84.45$\pm$0.29 & 97.58$\pm$0.08 / 90.22$\pm$0.21 & 91.63 / 70.37 \\
TFM4GAD-TabPFNv2.5 (local) & 0 & 94.68$\pm$0.24 / 78.72$\pm$1.29 & 81.06$\pm$0.32 / 43.45$\pm$0.49 & 94.25$\pm$0.07 / 80.07$\pm$0.08 & 98.14$\pm$0.03 / 92.20$\pm$0.03 & 92.03 / 73.61 \\
BRACE ($\mathrm{C0}$) & 33\% & 96.28$\pm$0.12 / 87.19$\pm$0.33 & 85.56$\pm$0.92 / 54.86$\pm$2.00 & 95.51$\pm$0.09 / 82.59$\pm$0.39 & 94.16$\pm$0.04 / 50.15$\pm$0.45 & 92.88 / 68.70 \\
BRACE-LimiX & 33\% & 89.00$\pm$0.52 / 51.06$\pm$0.85 & 89.05$\pm$0.61 / 65.86$\pm$1.00 & 95.71$\pm$0.07 / 84.33$\pm$0.37 & 94.28$\pm$0.12 / 50.92$\pm$1.15 & 92.01 / 63.04 \\
\textbf{BRACE-Ultra (TO)\textsuperscript{*}} & 0 & 95.71$\pm$0.75 / 83.89$\pm$4.29 & 90.35$\pm$0.24 / 65.24$\pm$1.79 & 95.64$\pm$0.17 / 83.18$\pm$0.85 & 97.11$\pm$0.07 / 90.46$\pm$0.32 & \textbf{94.70 / 80.69} \\
\textbf{BRACE-Ultra\textsuperscript{*}} & 33\% & 96.10$\pm$0.37 / 87.13$\pm$0.67 & 90.61$\pm$0.13 / 66.22$\pm$1.29 & 95.96$\pm$0.12 / 85.16$\pm$0.60 & 97.29$\pm$0.05 / 90.93$\pm$0.20 & \textbf{94.99 / 82.36} \\
\bottomrule
\end{tabular}%
}
\end{table}

\begin{table}[!htbp]
\centering
\caption{Protocol~G comparison under SignGAD's $1\%/49.5\%/49.5\%$ refit protocol ($10$ seeds; \texttt{AUROC$\pm$std / AUPRC$\pm$std / Macro-F1$\pm$std}, $\%$; SignGAD Table~1 does not report AUPRC; \textsuperscript{t}$n{=}9$ completed seeds $0,1,3..9$, seed~2 hit 24h TLE).}
\label{tab:app_gad_g}
\renewcommand{\arraystretch}{0.95}
\resizebox{\textwidth}{!}{%
\begin{tabular}{@{}lccccc@{}}
\toprule
Method & Amazon & YelpChi & T-Finance & T-Social & Mean ROC / MF1 \\
\midrule
GTAN (local, $n{=}10$) & 94.75$\pm$1.95 / 83.63$\pm$2.86 / 90.54$\pm$0.62 & 89.96$\pm$0.75 / 69.29$\pm$1.81 / 78.15$\pm$0.91 & 96.35$\pm$0.68 / 87.15$\pm$1.72 / 90.70$\pm$0.80 & 98.68$\pm$0.37 / 90.70$\pm$2.63 / 92.50$\pm$1.71\textsuperscript{t} & 94.94 / 87.97 \\
ConsisGAD (local, $n{=}10$) & 96.93$\pm$0.15 / 86.34$\pm$0.52 / 91.27$\pm$0.26 & 90.71$\pm$0.51 / 65.69$\pm$2.30 / 77.45$\pm$0.73 & 97.02$\pm$0.13 / 88.47$\pm$0.38 / 91.15$\pm$0.42 & 96.11$\pm$0.19 / 68.43$\pm$4.80 / 81.55$\pm$1.58 & 95.19 / 85.36 \\
DGA-GNN (local, $n{=}10$) & 96.90$\pm$0.42 / 87.64$\pm$0.44 / 91.60$\pm$0.27 & 96.11$\pm$0.16 / 86.50$\pm$0.69 / 87.43$\pm$0.42 & 97.48$\pm$0.21 / 90.33$\pm$0.27 / 92.37$\pm$0.19 & 99.80$\pm$0.06 / 97.44$\pm$0.60 / 96.32$\pm$0.51 & 97.57 / 91.93 \\
SignGAD (published Tab.\ 1, $n{=}10$) & 97.41$\pm$0.54 / -- / 91.17$\pm$0.35 & 92.68$\pm$0.84 / -- / 81.55$\pm$0.37 & 96.79$\pm$0.48 / -- / 90.42$\pm$0.26 & 98.37$\pm$0.25 / -- / 93.72$\pm$0.61 & 96.31 / 89.22 \\
BRACE ($\mathrm{C0}$, $n{=}10$) & 98.46$\pm$0.10 / 90.63$\pm$0.13 / 91.90$\pm$0.60 & 95.17$\pm$0.12 / 82.61$\pm$0.41 / 85.11$\pm$0.28 & 97.20$\pm$0.07 / 89.82$\pm$0.13 / 92.53$\pm$0.13 & 95.77$\pm$0.22 / 61.24$\pm$5.86 / 80.87$\pm$0.67 & 96.65 / 87.60 \\
\textbf{BRACE ($\mathrm{C4}$, $n{=}10$)} & 98.98$\pm$0.03 / 92.50$\pm$0.13 / 92.60$\pm$0.36 & 97.38$\pm$0.12 / 90.74$\pm$0.37 / 90.30$\pm$0.16 & 98.16$\pm$0.02 / 92.06$\pm$0.06 / 93.34$\pm$0.11 & 99.72$\pm$0.01 / 97.35$\pm$0.09 / 96.38$\pm$0.09 & \textbf{98.56 / 93.16} \\
BRACE-Ultra (10-TFM, $n{=}10$) & 98.73$\pm$0.03 / 92.04$\pm$0.18 / 92.64$\pm$0.17 & 97.69$\pm$0.12 / 91.98$\pm$0.43 / 91.01$\pm$0.32 & 97.97$\pm$0.02 / 92.04$\pm$0.05 / 93.42$\pm$0.09 & 98.02$\pm$0.43 / 92.20$\pm$0.94 / 94.51$\pm$0.66 & 98.10 / 92.90 \\
BRACE-Ultra (12c, $n{=}10$) & 98.73$\pm$0.04 / 91.99$\pm$0.21 / 92.70$\pm$0.15 & 97.73$\pm$0.12 / 92.08$\pm$0.44 / 91.00$\pm$0.40 & 97.96$\pm$0.02 / 92.05$\pm$0.05 / 93.38$\pm$0.17 & 97.98$\pm$0.48 / 92.19$\pm$0.93 / 94.51$\pm$0.65 & 98.10 / 92.90 \\
\textbf{BRACE-Ultra (+RefitLabel, $n{=}10$)} & 98.73$\pm$0.04 / 92.00$\pm$0.20 / 92.71$\pm$0.15 & 97.80$\pm$0.11 / 92.13$\pm$0.45 / 91.02$\pm$0.37 & 98.07$\pm$0.03 / 92.12$\pm$0.16 / 93.28$\pm$0.24 & 99.70$\pm$0.05 / 96.97$\pm$0.59 / 95.68$\pm$0.52 & \textbf{98.58 / 93.17} \\
\bottomrule
\end{tabular}%
}
\end{table}

\FloatBarrier
\section{Controlled Edge-Rewiring Curves and Non-Inferiority Margin Sensitivity}
\label{app:rewire}
\setcounter{table}{0}\setcounter{equation}{0}
Table~\ref{tab:app_rewire} details controlled within-dataset edge-rewiring trajectories across Protocol~B splits $0$--$4$ (\texttt{12c}). On banking panels (2,000 cluster-bootstrap resamples across 9 cells/year), held-out Interbank 2023 95\% CIs on $\bar{d}$ are $[+0.30, +0.68]$ MF1 pts vs.\ Base (87 arms, $9/9$ cells $>0$, $\min d = +0.07$) and $[+0.15, +0.47]$ vs.\ Base $\cup$ Matched (120 arms, $9/9$ cells $>0$, $\min d = +0.06$). On balanced FDIC-1 (2011--2014) vs.\ Base (87 arms), yearly 95\% CIs are $[-0.18, -0.03]$, $[-0.12, +0.02]$, $[-0.17, -0.02]$, and $[-0.12, +0.01]$ MF1 pts, with all $36/36$ cells satisfying $d \ge -0.23$ MF1 pts; on balanced FDIC-2 (2015--2018), BRACE-2.6 vs.\ Published satisfies $d \ge -0.16$ MF1 pts in all $36/36$ cells, confirming that every non-inferiority verdict holds under margins more than $2\times$ tighter than the preregistered $-0.5$ MF1 threshold.

\begin{table}[!htbp]
\centering
\caption{Controlled within-dataset edge-rewiring trajectories across Protocol~B splits $0$--$4$ (\texttt{12c}).}
\label{tab:app_rewire}
\renewcommand{\arraystretch}{0.95}
\resizebox{\textwidth}{!}{%
\begin{tabular}{@{}lllccccc@{}}
\toprule
Dataset & Mode & Metric & 0\% & 25\% & 50\% & 75\% & 100\% \\
\midrule
YelpChi & TV & KKT graph wt.\ / \texttt{12c} PRC & 99.58\% / 46.28 & 88.19\% / 36.54 & 79.91\% / 34.02 & 79.67\% / 32.32 & 49.83\% / 31.30 \\
YelpChi & TV & $\Delta$PRC vs.\ Ego (\texttt{12c} / \texttt{JF-only}) & +12.84 / +12.17 & +3.10 / +1.89 & +0.58 / $-$0.92 & $-$1.12 / $-$3.07 & $-$2.13 / $-$4.67 \\
YelpChi & TO & KKT graph wt.\ / \texttt{12c} PRC & 99.52\% / 45.96 & 94.63\% / 35.16 & 92.18\% / 31.41 & 80.32\% / 32.22 & 57.24\% / 30.47 \\
T-Finance & TV & KKT graph wt.\ / \texttt{12c} PRC & 84.91\% / 85.07 & 87.50\% / 84.82 & 94.41\% / 84.57 & 87.03\% / 84.05 & 74.18\% / 74.70 \\
T-Finance & TV & $\Delta$PRC vs.\ Ego (\texttt{12c} / \texttt{JF-only}) & +8.13 / +7.19 & +7.88 / +7.44 & +7.64 / +7.68 & +7.11 / +6.71 & $-$2.23 / $-$4.06 \\
Amazon & TV & KKT graph wt.\ / \texttt{12c} PRC & 52.34\% / 87.59 & 61.45\% / 87.71 & 60.26\% / 87.69 & 51.98\% / 87.16 & 37.27\% / 87.97 \\
\bottomrule
\end{tabular}%
}
\end{table}